\documentclass[conference]{IEEEtran}

\IEEEoverridecommandlockouts

\usepackage{amsmath,amssymb,amsfonts}
\usepackage{booktabs}
\usepackage[font=small,labelfont=bf]{caption}
\usepackage{subcaption}
\usepackage{graphicx}
\usepackage{textcomp}
\usepackage{xcolor}
\usepackage{url}
\usepackage{hyperref}
\hypersetup{colorlinks=true, linkcolor=black, citecolor=blue, urlcolor=blue}
\usepackage{fancyhdr}
\usepackage{placeins}     

\fancypagestyle{plain}{%
  \fancyhf{}%
  \fancyfoot[C]{\footnotesize\copyright{} 2026 JP Morgan Chase \& Co. All
                rights reserved \quad\thepage}%
}

\def\BibTeX{{\rm B\kern-.05em{\sc i\kern-.025em b}\kern-.08em
    T\kern-.1667em\lower.7ex\hbox{E}\kern-.125emX}}

\newcommand{\R}{\mathbb{R}}
\newcommand{\E}{\mathbb{E}}
\newcommand{\norm}[1]{\left\lVert#1\right\rVert}
\newcommand{\inner}[2]{\left\langle#1,\,#2\right\rangle}
\newcommand{\T}{\top}

\begin{document}

\title{Representation Without Reward:\\
  A JEPA Audit for LLM Fine-Tuning}

\author{%
\IEEEauthorblockN{Biswa Sengupta}
\IEEEauthorblockA{%
  LLM Suite Team, JPMorgan Chase \& Co.\\
  Email: biswa.sengupta@jpmorgan.com}
}

\maketitle

\begin{abstract}
Joint-embedding predictive architectures (JEPAs) propose that a model
should learn more useful abstractions when trained to predict latent
representations rather than observed outputs.  For autoregressive
language-model fine-tuning the principle entails a stricter requirement:
the induced hidden-state geometry must reach the language-model head
\emph{and} improve the decoded task metric.  We test that requirement
under a fixed Llama-3.2-1B-Instruct LoRA harness on
natural-language-to-regex generation, comparing twenty-two training-time
auxiliaries across trajectory-shape regularisation, distributional
constraints, predictor/target asymmetry, Fisher-metric Jacobi residuals,
and a decoder-visible JEPA objective constructed to lie in
cross-entropy's positive cone.  The empirical answer is a structured
null: several auxiliaries clear single-cell paired $\alpha = 0.10$
without correction (T3-Local at $\Delta = +2.53$~pp, $p = 0.003$
being the strongest), but none survives Bonferroni or
Holm--Bonferroni at the relevant family-wise threshold, even though
many change curvature, anisotropy, variance, and gradient direction.  Decoder-visible JEPA
yields the first positive auxiliary--cross-entropy gradient cosine in
the study, yet exact match remains inside seed noise; a full-fine-tuning
replication of the same auxiliary at $n = 5$ seeds reproduces the null
on both benchmarks (TURK: $\Delta = +0.04$~pp, $p_{\text{paired}} = 0.96$;
SYNTH: $\Delta = +0.52$~pp, $p_{\text{paired}} = 0.28$), so the null
is robust across LoRA and full fine-tuning for the decoder-visible
construction.  Hidden-state representation work and decoded-task
accuracy are therefore weakly coupled in this regime; we accordingly
reframe LLM-domain JEPA evaluation as a coupling problem, in which the
operative question is under which metrics useful hidden geometry
becomes decoder-visible task signal.
\end{abstract}

\medskip
\noindent{\footnotesize\itshape\textbf{Disclaimer:} This paper was
prepared for informational purposes by the LLM Suite group of JP
Morgan Chase and its affiliates (`JPMC') and is not a product of
the Research Department of JP Morgan. JP Morgan makes no
representation, warranty or undertaking whatsoever and disclaims
all liability for the completeness, accuracy or reliability of the
information contained herein. This document is not intended as
investment research or investment advice, or a recommendation,
offer or solicitation for the purchase or sale of any security,
financial instrument, financial product or service, or to be used
in any way for evaluating the merits of participating in any
transaction, and shall not constitute a solicitation under any
jurisdiction or to any person, if such solicitation under such
jurisdiction or to such person would be unlawful.\par}\medskip

\section{Introduction}

Joint-embedding predictive architectures propose a compelling learning
principle: predict the representation of what is missing, not the
pixels or tokens themselves. Variants of this idea appear across
several research programmes:
self-supervised predictive networks built around prediction
error~\cite{schmidhuber1990}, the joint-embedding predictive
architecture (JEPA) framing that names the canonical
form~\cite{lecun2022jepa}, and curiosity-driven model-building that
extends the prediction-error idea to
exploration~\cite{schmidhuber1991curious}. The principle is currently
exemplified in vision by I-JEPA~\cite{assran2023ijepa} and dense video
extensions~\cite{murlabadia2026vjepa21}, which combine latent
prediction with predictor/target asymmetry. The scientific question for language-model
fine-tuning is not whether such objectives can move hidden states. They
can. The sharper question is whether the movement is coupled to the
decoder and to the task metric.

Autoregressive language models make this question unusually stringent.
The training signal is next-token cross-entropy, the downstream system is
the language-model head plus a decoding rule, and many symbolic tasks are
scored by exact match. A representation may become smoother, more
isotropic, or more predictive under an auxiliary loss while leaving the
selected token sequence unchanged. We therefore separate two testable
claims that the JEPA principle conflates in this setting:
\begin{quote}
\emph{H1 (representation activity).} Adding a JEPA-style auxiliary to
LoRA fine-tuning induces a measurable change in hidden-state geometry
relative to the no-auxiliary baseline --- in anisotropy, trajectory
curvature, seed-to-seed variance, or gradient direction.\\[2pt]
\emph{H2 (decoder-coupled payoff).} That hidden-state change
translates, on the same checkpoints, into a measurable improvement of
the decoded exact-match accuracy.
\end{quote}
H1 is about whether the auxiliary \emph{acts}; H2 is about whether the
exact-match metric \emph{rewards} the action.  The two claims are not
redundant: an auxiliary that satisfies H1 but not H2 is the empirical
signature this paper sets out to characterise.

Semantic Tube Prediction (STP)~\cite{huang2026stp} provides a minimal
language-domain version of the JEPA idea. It treats hidden states along
an assistant response as a local trajectory on a semantic manifold and
adds a cosine-alignment loss between velocity vectors. The reported
$16\times$ data-efficiency gain on NL-RX-SYNTH motivates a broader
hypothesis: perhaps trajectory- or representation-predictive pressure can
supply task-relevant structure when only LoRA adapters are being trained.

We test that hypothesis by fixing the model, LoRA setup, data format,
splice point, decoder, and exact-match evaluator, and varying only the
auxiliary objective. The study covers twenty-two training-time
auxiliaries and one inference-time projector for Llama-3.2-1B-Instruct
on natural-language-to-regex generation. The objectives are organised not
as an inventory of losses, but as a sequence of mechanistic questions:
Do trajectory attractors help? Do contrastive or predictive pressures
help? Do distributional constraints outside cross-entropy's
implicit-bias direction help? Is JEPA's predictor/target asymmetry the
missing ingredient? If hidden geometry matters but the Euclidean metric
is wrong, does a Fisher-metric Jacobi residual expose it? If gradient
misalignment is the obstacle, can a decoder-visible JEPA objective force
the auxiliary into cross-entropy's positive cone?

The answer across these tests is a structured null. Without
statistical correction, several auxiliaries reach single-cell
paired $\alpha = 0.10$: prompt-local JFR on TURK ($+2.53$~pp,
$p = 0.003$), T5 on TURK ($+1.53$~pp, $p = 0.08$), and three
Tier-1 distributional cells on SYNTH (L1, L3, L4 with paired $p$
between $0.06$ and $0.10$). With Bonferroni or Holm--Bonferroni
correction at the relevant family-wise threshold, none of these
cells survives. BYOL-LLM, the direct test of predictor + EMA
target + stop-gradient asymmetry, remains inside seed noise on
both benchmarks. The Fisher-metric replacements remove the
geometric signature without improving exact match. Decoder-visible
JEPA produces the first positive auxiliary--cross-entropy gradient
cosine, but its exact-match gain also remains statistically
unresolved both under LoRA and under a full-fine-tuning replication
at $n = 5$ seeds.

The null is therefore not a story of inactive objectives. Several losses
reduce trajectory curvature, alter hidden-state anisotropy, tighten
seed-to-seed variance, or change gradient direction. The paper's central
claim is instead a separation between representation work and decoded
performance: in this harness, JEPA-style auxiliaries can reshape hidden
geometry without reliably changing the exact symbolic output selected by
the language-model head.

\paragraph{Contributions.}
\begin{enumerate}
\item \textbf{A decoder-visible test of JEPA-style representation
  shaping in LLM fine-tuning.} We frame the central question as whether
  auxiliary representation objectives merely move hidden geometry or
  produce changes that survive the language-model head and improve
  exact-match decoding.

\item \textbf{A controlled hypothesis map.} We evaluate
  twenty-two JEPA-inspired training-time auxiliaries and one
  inference-time projector on Llama-3.2-1B-Instruct, keeping the model,
  LoRA setup, data, splice point, decoder, and evaluator fixed. The
  menu covers trajectory-shape attractors (STP, T1--T6), contrastive
  and predictive objectives (T7, L12), distributional constraints
  outside cross-entropy's implicit-bias direction (L1--L6, L9),
  predictor/target asymmetry tests (L13--L14), Fisher-metric Jacobi
  residuals (Tier-3), and a decoder-visible JEPA objective constructed
  to lie in cross-entropy's positive cone.

\item \textbf{A structured null for asymptotic exact-match accuracy.}
  Without statistical correction, a handful of cells reach single-cell
  paired $\alpha = 0.10$ (prompt-local JFR and T5 on TURK; L1, L3, L4
  on SYNTH), but none survives Bonferroni or Holm--Bonferroni at the
  relevant family-wise threshold. This includes BYOL-LLM (predictor +
  EMA target + stop-gradient asymmetry) and decoder-visible JEPA
  (gradient inside the cross-entropy positive cone), both of which
  remain inside seed noise.

\item \textbf{Diagnostics separating performance inactivity from
  representational inactivity.} Several auxiliaries alter curvature,
  anisotropy, variance, and gradient direction even when exact-match
  accuracy is unchanged. The losses act; their action is weakly coupled
  to the decoder and metric in this harness.
\end{enumerate}

\paragraph{Paper organisation.}
Section~\ref{sec:related} situates the paper against the JEPA, contrastive
and distributional self-supervision literature.
Section~\ref{sec:methods} introduces the unified splice point and gives
formal definitions for every auxiliary in three structural classes.
Section~\ref{sec:setup} describes the experimental harness and the
statistical methodology, including the Welch's-test and family-level
corrections we use throughout. Section~\ref{sec:results} reports
TURK and SYNTH results, the within-family Jacobi-field analysis, and the
two failure modes (T1 collapse, T9 fixed-point decoding).
Section~\ref{sec:discussion} synthesises the diagnostics with the
benchmark results into the decoder-invisible representation-work reading
and discusses implications for future LLM-domain JEPA work.

\section{Related Work}\label{sec:related}

\paragraph{Joint-embedding predictive architectures.}
JEPA methods learn by predicting representations rather than reconstructing
inputs.  I-JEPA~\cite{assran2023ijepa} established this template for images
with masked prediction, a learned predictor, and an EMA target encoder;
V-JEPA-style extensions~\cite{murlabadia2026vjepa21} extend the same idea to
video and dense intermediate features.  These systems typically rely on
asymmetry---for example target encoders, stop-gradient operations, or
predictor heads---to prevent collapse while preserving semantic information.
Our study intentionally removes most of this architecture in order to test a
minimal question: how much can a single-forward-pass auxiliary loss attached
to an LLM fine-tuning loop accomplish by itself?

\paragraph{Contrastive, non-contrastive, and distributional SSL.}
Contrastive objectives such as SimCLR~\cite{simclr2020} use negative pairs to
shape representation geometry, while non-contrastive methods such as
BYOL~\cite{byol2020}, SimSiam~\cite{simsiam2021}, VICReg~\cite{vicreg2022},
and Barlow Twins~\cite{barlowtwins2021} use predictor asymmetry, variance,
or covariance constraints.  Recent theory connects these families through
implicit variance regularisation and dualities between contrastive and
non-contrastive objectives~\cite{tian2021byolworks,garrido2024duality,halvagal2023}.
These results motivate the distributional losses in our study: if STP-style
trajectory alignment fails because it is too narrow, isotropy or covariance
regularisation might still move the representation in a useful direction.

\paragraph{JEPA theory and language-domain variants.}
Recent analyses argue that JEPA-style objectives can prefer semantically
useful features under appropriate depth or architectural asymmetry
\cite{littwin2024jepa}, and that isotropic Gaussian embedding distributions
can arise from minimax prediction-risk formulations~\cite{lejepa2025}.
Language-domain variants include data2vec~\cite{data2vec2022}, LLM-JEPA
\cite{huang2025llmjepa}, and STP~\cite{huang2026stp}.  LLM-JEPA uses two
views and a predictor/target-style mechanism; STP strips the idea down to a
single causal forward pass over the assistant trajectory.  Our paper is best
read as a controlled study of that stripped-down regime.  We do not test whether the
full JEPA recipe with target encoders and predictor asymmetry can help LLMs;
we test whether trajectory and distributional pressures alone improve exact
match when added to LoRA fine-tuning.

\paragraph{Trajectory straightening as a regulariser.}
The straightening principle that motivates STP and our T1--T6 cells has
recently been shown to lift task performance in a different downstream
setting. \cite{wang2026temporalstraightening} train a world-model encoder
plus predictor with a curvature regulariser that is structurally
identical to STP --- a cosine penalty between consecutive latent
velocities --- and report a $20$--$60\%$ improvement in goal-reaching
success under gradient-based latent planning. They additionally prove
that under a linear-dynamics assumption the straightening regulariser
controls the condition number of the planning Hessian, providing a
direct mechanism for the empirical gain. Their result is not in tension
with our null: their downstream task is gradient-based planning over a
learned latent dynamics model, where representation conditioning enters
the optimisation Hessian directly; ours is autoregressive token
generation under cross-entropy followed by argmax decoding, where the
representation enters only through the LM head's softmax. The
contrast sharpens our structured-null reading
(Section~\ref{sec:discussion:parsimonious}): the same geometric work
that improves a gradient-based downstream objective need not improve a
metric whose dependence on the representation is mediated only by the
final-token argmax.  The decoder-visible JEPA construction we pursue
in Appendix~\ref{appx:dvjepa} addresses this asymmetry directly by
moving the auxiliary out of $h$-space and into the post-softmax
distribution of the LM head.

\paragraph{Positioning.}
The closest comparison is STP~\cite{huang2026stp}, which reports a
$16\times$ data-efficiency gain on NL-RX-SYNTH.  Our evaluation differs in
scope: we measure final exact-match accuracy after LoRA fine-tuning, include
NL-RX-TURK as a less saturated benchmark, and add diagnostics for curvature,
anisotropy, and gradient alignment.  The two claims are therefore compatible:
a geometric auxiliary may improve sample efficiency without improving the
asymptotic exact-match ceiling.  Our contribution is to show that, under this
controlled exact-match harness, the representational changes are real but not
reliably task-improving.

\section{Methods}\label{sec:methods}

We adopt a single notation throughout. Let $H \in \R^{B \times S
\times D}$ denote the final-layer hidden states of a batch of
$B$ training examples, with $S$ the padded sequence length and
$D = 2048$ the hidden dimension of Llama-3.2-1B. For each
example $b$, let $[\text{lo}_b,\,\text{hi}_b)$ be the half-open
range of token positions occupied by the assistant turn (the regex
output), and let $L_b = \text{hi}_b - \text{lo}_b$ be its length.

The total training loss is
\begin{equation}
  \mathcal{L}_{\text{total}} \;=\; \mathcal{L}_{\text{LM}}
  \;+\; \lambda(t)\,\mathcal{L}_{\text{aux}},
  \label{eq:total}
\end{equation}
where $\mathcal{L}_{\text{LM}}$ is the standard token-level
cross-entropy on the assistant span (and the trailing
end-of-turn token), $\lambda(t)$
is the auxiliary weight at training step $t$ (under our
warmup-decay schedule, see Section~\ref{sec:setup}), and
$\mathcal{L}_{\text{aux}}$ is one of the auxiliaries defined
below. The single inference-time intervention, T9 Tube-Projected
Decoding, is described separately in
Section~\ref{sec:methods:t9}; it does not enter
$\mathcal{L}_{\text{total}}$.

The experimental template can be written as a single template:
\begin{equation}
  \mathcal{L}_{\text{aux}}
  \in
  \left\{
  \mathcal{L}_{\text{traj}},\,
  \mathcal{L}_{\text{dist}},\,
  \mathcal{L}_{\text{pred}}
  \right\},
  \label{eq:aux-template}
\end{equation}
where
\begin{align}
  \mathcal{L}_{\text{traj}}
  &= \E_{b,t}\!\left[D_{\text{geom}}(h_{b,t-1},h_{b,t},h_{b,t+1})\right],
  \label{eq:traj-template}\\
  \mathcal{L}_{\text{dist}}
  &= D_{\mathcal{P}}\!\left(\widehat P_{\psi(H_{\text{span}})}, P_0\right),
  \label{eq:dist-template}\\
  \mathcal{L}_{\text{pred}}
  &= d\!\left(q_\phi(z_{\text{context}}),\operatorname{sg}(z_{\text{target}})\right).
  \label{eq:pred-template}
\end{align}
The individual losses below are instantiations of these three forms.  This
notation is useful because it separates what is being tested: trajectory
shape, feature-distribution shape, or context-target prediction.

\begin{table}[t]
\centering
\caption{Conceptual design axes.  The study is intended to falsify mechanisms,
not merely to enumerate losses.}
\label{tab:design-axes}
\small
\setlength{\tabcolsep}{3pt}
\begin{tabular}{p{0.31\columnwidth}p{0.21\columnwidth}p{0.34\columnwidth}}
\toprule
Question & Cells & Mechanism tested\\
\midrule
Does STP-style straightening help? & STP, T1--T6 & Geometry-only tube regularisation\\
Does negative or temporal prediction help? & T7, L12 & Discriminative / predictive pressure\\
Does escaping CE implicit bias help? & L1--L6, L9 & Marginal or tangent distribution shaping\\
Is JEPA asymmetry sufficient? & L13--L14 & Predictor + target + stop-gradient\\
Does a task-aware metric help? & Fisher-JFR, Fisher-MSTB, Fisher-Local-JFR & Replace Euclidean tube with the LM-head Fisher pull-back (decoder-aligned geometry)\\
Does decoder-visible JEPA help? & DV-JEPA & Move auxiliary out of $h$-space into the post-softmax distribution of the LM head\\
\bottomrule
\end{tabular}
\end{table}

\subsection{Where the Auxiliary Loss Attaches}\label{sec:methods:integration}

\begin{figure}[t]
\centering
\includegraphics[width=\columnwidth]{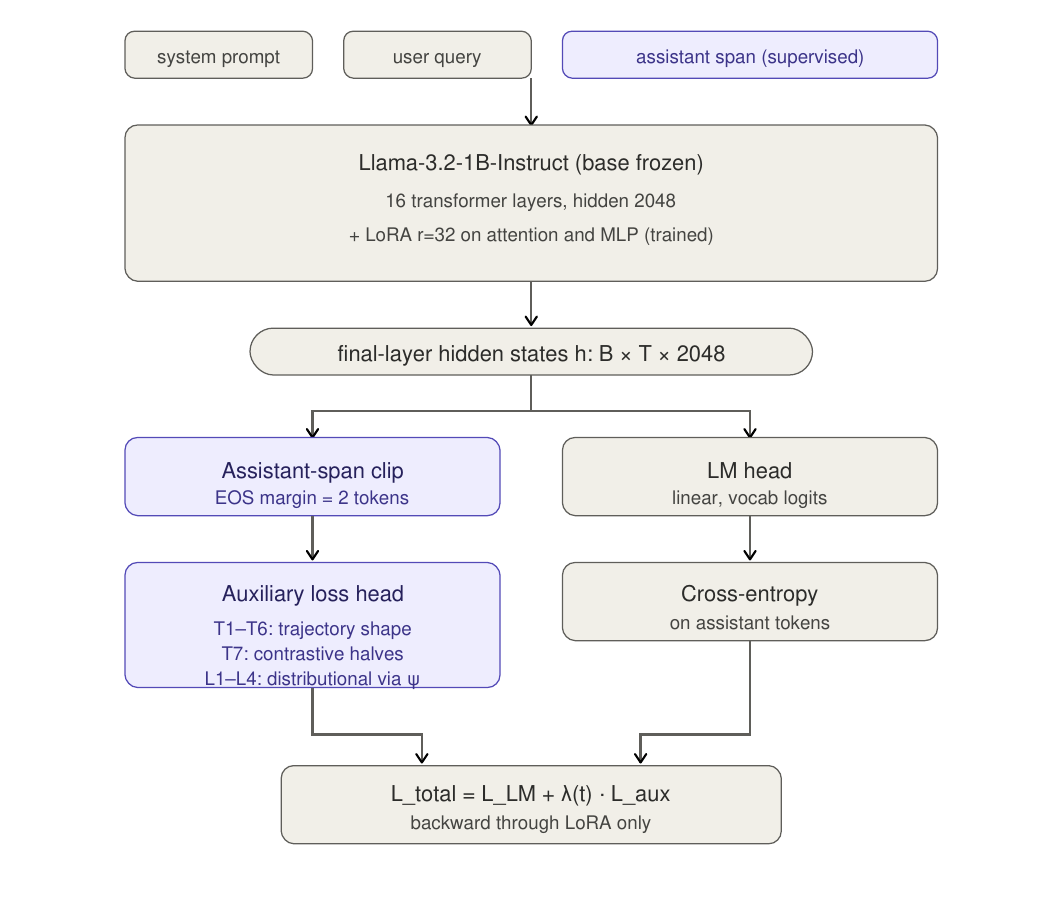}
\caption{The auxiliary loss splices into the standard LoRA
fine-tuning pipeline at the final-layer hidden states, in
parallel with the language-model head.}
\label{fig:aux-integration}
\end{figure}

Every auxiliary in this study shares a single splice point in
the forward pass, illustrated in
Fig.~\ref{fig:aux-integration}. The base model returns the
final-layer hidden states
$h \in \R^{B \times T \times D}$ from a single forward pass;
the auxiliary head and the language-model head consume the
same tensor in parallel, with no second backbone pass and no
modification to the transformer architecture itself.

The LM-head branch is unchanged: it projects every position to
vocabulary logits and computes the cross-entropy on the
assistant span plus the end-of-turn token.  The auxiliary
branch first applies the EOS clip
(Section~\ref{sec:setup}) and then dispatches to one of three
structural families --- trajectory-shape (STP, T1, T2, T3,
T3-Local, T5, T6), in-batch contrastive (T7), and distributional
(L1--L4 projected through a sketcher
$\psi : \R^D \to \R^{d'}$).  The two scalar losses combine
as in~\eqref{eq:total}; the LoRA adapters and the auxiliary
head's parameters carry the gradient, and the base Llama
weights stay frozen.  T5 (Deep Semantic Tubes) is the lone
variant that additionally reads intermediate layers
$\{h^{(4)}, h^{(8)}, h^{(12)}, h^{(16)}\}$; every other
variant operates only on the final-layer state.

Table~\ref{tab:interventions} lists the eighteen training-time
auxiliaries plus the single inference-time intervention; the
four falsification-route additions (Tier-3 Fisher variants and
decoder-visible JEPA) are documented in
Appendices~\ref{appx:fisher} and~\ref{appx:dvjepa}.

\begin{table}[t]
\centering
\caption{The hypothesis map: eighteen training-time auxiliaries plus
one inference-time intervention.}
\label{tab:interventions}
\small
\setlength{\tabcolsep}{3pt}
\begin{tabular}{llcc}
\toprule
\# & Name & Class & Eq.\\
\midrule
STP & Semantic Tube Prediction & 1st-order attractor   & \eqref{eq:stp}\\
T1  & Curvature-Aware Tube     & 2nd-order attractor   & \eqref{eq:ctube}\\
T2  & Riemannian-Metric Tube   & metric-cosine attractor & \eqref{eq:rig}\\
T3  & Jacobi-Field Regulariser & 2nd-order attractor   & \eqref{eq:jfr}\\
T3-Local & Prompt-Local JFR     & T3 with memory bank   & \eqref{eq:jfrlocal}\\
T5  & Deep Semantic Tubes      & T3 across layers      & \eqref{eq:dst}\\
T6  & Multi-Scale Tube Bundle  & T3 across scales      & \eqref{eq:mstb}\\
T7  & Contrastive Tube         & in-batch InfoNCE      & \eqref{eq:t7}\\
T9  & Tube-Projected Decoding  & inference-time        & \eqref{eq:tpd}\\
\midrule
L1  & SIGReg-State             & state isotropy        & \eqref{eq:sigregstate}\\
L2  & SIGReg-Tangent           & tangent isotropy      & \eqref{eq:sigregtangent}\\
L3  & C-Tube-Sectional         & curvature-variance    & \eqref{eq:sectional}\\
L4  & STP-CMF                  & half-vs-half CF-MMD   & \eqref{eq:stpcmf}\\
\midrule
L5  & VICReg-VC                & variance + covariance & \eqref{eq:vicregvc}\\
L6  & SW-Iso                   & sliced-Wasserstein    & \eqref{eq:swiso}\\
L9  & Score-Match              & Hyvärinen + N(0,I) score & \eqref{eq:scorematch}\\
L12 & CPC                      & temporal InfoNCE      & \eqref{eq:cpc}\\
L13 & BYOL-LLM                 & EMA target + predictor & \eqref{eq:byol}\\
L14 & I-JEPA-LLM               & masked-block predict  & \eqref{eq:ijepa}\\
\bottomrule
\end{tabular}
\end{table}

Three conceptual classes organise the discussion that follows.
\emph{Trajectory-shape auxiliaries} (STP, T1--T6) penalise
directional misalignment, curvature, or higher-order differences
across the assistant span, drawing on the geodesic intuition
behind STP~\cite{huang2026stp}.  \emph{Contrastive and
predictive auxiliaries} (T7~\cite{simclr2020}, L12) introduce
discriminative structure, either across span halves or across
positions.  \emph{Distributional and predictor-based
auxiliaries} (L1--L6, L9, L13, L14) project hidden states or
tangents through a sketcher and either enforce isotropy
(via Cram\'er--Wold CF distance, sliced-Wasserstein, or score
matching) or learn a predictor against an EMA / fixed target
that probes whether CE's implicit-bias
direction~\cite{soudry2018} can be supplemented by an
anti-collapse prior.

\subsection{STP -- First-Order Direction Alignment}
For each example we sample four indices $i_1 < i_2 < i_3 < i_4$
inside the assistant span and form
\begin{equation}
  u_b = h_{b,i_2} - h_{b,i_1}, \qquad v_b = h_{b,i_4} - h_{b,i_3}.
\end{equation}
The STP loss is the directional misalignment of $u$ and $v$:
\begin{equation}
  \mathcal{L}_{\text{STP}}
  \;=\; \frac{1}{B}\sum_b
  \left[ 1 - \frac{\inner{u_b}{v_b}}{\norm{u_b}\,\norm{v_b}} \right].
  \label{eq:stp}
\end{equation}
Because both vectors are unit-normalised, $\mathcal{L}_{\text{STP}}$
depends only on the angle between two velocity vectors of the
trajectory; its gradient is perpendicular to each velocity and
vanishes when the trajectory traces a fixed-direction ray.

\subsection{T1 -- Curvature-Aware Tube (C-Tube)}
Given four indices $s < p < q < t$, define the chord
$\tau_b = h_{b,t} - h_{b,s}$ and the average central second
difference
\begin{equation}
\begin{aligned}
  d_b^{(2)} \;=\; & \tfrac{1}{2}\bigl[(h_{b,q} - h_{b,p}) - (h_{b,p} - h_{b,s})\bigr] \\
  & {} + \tfrac{1}{2}\bigl[(h_{b,t} - h_{b,q}) - (h_{b,q} - h_{b,p})\bigr].
\end{aligned}
\end{equation}
Let $\kappa_b = d_b^{(2)} - \frac{\inner{d_b^{(2)}}{\tau_b}}{\norm{\tau_b}^2}\tau_b$ be its component orthogonal to the chord. Then
\begin{equation}
  \mathcal{L}_{\text{T1}}
  \;=\; \frac{1}{B}\sum_b \norm{\kappa_b}_2^2.
  \label{eq:ctube}
\end{equation}
$\mathcal{L}_{\text{T1}} = 0$ exactly when the trajectory is a
discrete geodesic (zero geodesic curvature). For a $C^2$ trajectory
sampled at spacing $\Delta$, the second-difference estimator
$d_b^{(2)}$ is consistent for the geodesic curvature with relative
bias $\mathcal{O}(\Delta^2)$, so the loss value $\norm{\kappa_b}^2$
scales as $\Delta^4 \norm{\kappa_{\text{geo}}}^2 + \mathcal{O}(\Delta^6)$.
The corresponding STP loss scales as $\Delta^2 \kappa_{\text{geo}}^2$,
so on a fixed-curvature trajectory $\mathcal{L}_{\text{T1}} /
\mathcal{L}_{\text{STP}}$ vanishes as $\mathcal{O}(\Delta^2)$ as
$\Delta \to 0$.

\subsection{T2 -- Riemannian-Metric Tube (RIG-Tube)}
We replace the Euclidean cosine in~\eqref{eq:stp} with a cosine
under a learned Riemannian metric $g_\phi(h)$. Concretely,
$g_\phi(h) = I + U(h) U(h)^\T + \mathrm{diag}(\exp d(h))$
is a low-rank-plus-diagonal symmetric-positive-definite (SPD)
perturbation of the identity, parameterised by a small
multi-layer perceptron (MLP) head $\phi$ that maps a hidden state
$h$ to a rank-$r$ factor $U(h) \in \R^{D \times r}$ and a
log-diagonal $d(h) \in \R^D$.  The three additive terms each play
a distinct role: \emph{(i)} the ambient identity $I$ guarantees the
metric is at least the Euclidean inner product, so $g_\phi(h) \succeq
I \succ 0$ holds by construction with no extra constraint; it also
recovers plain STP at initialisation when $U \equiv 0$ and $d
\equiv -\infty$.  \emph{(ii)} The low-rank correction
$U(h) U(h)^\T$ concentrates curvature along a learned $r$-dimensional
subspace --- the directions in which the model is allowed to lengthen
or shorten its tangent inner products without the cost of a full
$D \times D$ parameterisation.  \emph{(iii)} The positive diagonal
$\mathrm{diag}(\exp d(h))$ supplies a per-axis scale that does not
have to fit inside the rank-$r$ subspace; the exponential
parameterisation enforces strict positivity of every diagonal entry.
Their sum is positive definite as a sum of positive-definite terms,
which avoids any need for projection or clipping during training. With $a = h_{t} - h_{r}$,
$b = h_{r} - h_{s}$ (three indices $s < r < t$), the loss is
\begin{equation}
  \mathcal{L}_{\text{T2}}
  \;=\; \E\!\left[
    1 - \frac{\inner{a}{b}_{g_\phi(h_r)}}
             {\norm{a}_{g_\phi(h_r)}\,\norm{b}_{g_\phi(h_r)}}
  \right],
  \label{eq:rig}
\end{equation}
where $\inner{x}{y}_g = x^\T g\, y$ and $\norm{x}_g = \sqrt{\inner{x}{x}_g}$.
We initialise $g_\phi \equiv I$ so that $\mathcal{L}_{\text{T2}}$ is
bit-identical to plain STP at step zero (curriculum guarantee:
adding T2 cannot hurt before training).

\subsection{T3 -- Jacobi-Field Regulariser (JFR)}
For a batch sharing a common prefix (or, in our NL-RX setting, no
shared prefix and the empty-prefix case) define the Jacobi-field
residual at position $t$ as
$J_b(t) = h_{b,t} - \bar{h}(t)$ where
$\bar{h}(t) = \tfrac{1}{B}\sum_b h_{b,t}$ is the batch
centroid trajectory. The flat-manifold limit of the Jacobi
equation requires $\ddot J_b \equiv 0$ along the trajectory, i.e.\
each residual trajectory is linear. Discretising with a central
3-point stencil,
\begin{equation}
  \mathcal{L}_{\text{T3}}
  \;=\; \frac{1}{|\mathcal{V}|}\!\!\sum_{(b,t)\,\in\,\mathcal{V}}\!\!
  \bigl\lVert J_b(t+1) - 2 J_b(t) + J_b(t-1) \bigr\rVert_2^2,
  \label{eq:jfr}
\end{equation}
where $\mathcal{V} = \{(b,t) : 1 \le t \le L_b - 2\}$ is the set
of interior token positions across the (variable-length) batch
and $|\mathcal{V}| = \sum_b (L_b - 2)$ is its cardinality. The
implementation realises $|\mathcal{V}|$ exactly as the sum of a
$0/1$ pad mask, so the normalisation matches the equation
bit-for-bit on variable-length batches.
$\mathcal{L}_{\text{T3}} = 0$ iff every residual trajectory is
exactly linear in $t$.

\subsection{T3-Local -- Prompt-Local JFR with Memory Bank}\label{sec:methods:t3local}
T3 \eqref{eq:jfr} subtracts the batch centroid before applying
the second-difference penalty, but in NL-RX where examples
have unrelated prompts the batch centroid is a noisy tube
centre. T3-Local replaces the batch mean by a
retrieval-weighted centroid drawn from a small memory bank of
past trajectories. Let
$\mathcal{M} = \{(a^{(c)}, H^{(c)}, L^{(c)})\}_{c=1}^{M}$
store, for the most recent $M$ training examples, the prompt
anchor $a^{(c)}$ (the user-message hidden-state mean), the
assistant trajectory $H^{(c)}$, and its length $L^{(c)}$. For
example $b$ with anchor $a_b$, take the top-$k$ neighbours by
cosine similarity and attention-weight their trajectories,
\begin{equation}
\bar h_b(t) = \sum_{j=1}^{k} w_{b,j}\,H^{(c_j)}(t), \;
w_{b,j} = \mathrm{softmax}_j\!\Big(\tfrac{\langle a_b,\,a^{(c_j)}\rangle}
                                          {\tau\,\lVert a_b\rVert\,\lVert a^{(c_j)}\rVert}\Big),
\end{equation}
and apply the T3 stencil to the residual
$J_b(t) = h_{b,t} - \mathrm{sg}[\bar h_b(t)]$:
\begin{equation}
\mathcal{L}_{\text{T3-Local}} = \frac{1}{|\mathcal{V}|}\!\!
  \sum_{(b,t)\in\mathcal{V}}\!\!
  \bigl\lVert J_b(t+1) - 2J_b(t) + J_b(t-1)\bigr\rVert_2^2.
  \label{eq:jfrlocal}
\end{equation}
The stop-gradient $\mathrm{sg}[\cdot]$ ensures the memory bank
acts as a target encoder, not a second backward path: the
auxiliary shapes the current trajectory toward the local
centroid of stored neighbours without pulling the stored
trajectories. Defaults: $k=8$, $\tau=0.1$, $M=512$. Conceptually
this is the closest variant we test to the EMA-target
asymmetry the JEPA literature identifies as essential
(Section~\ref{sec:discussion:diagnosis}).

\subsection{T5 -- Deep Semantic Tubes (DST-JFR)}\label{sec:methods:t5}
Apply the JFR loss at a fixed subset $\mathcal{L}$ of transformer
layers and average uniformly:
\begin{equation}
  \mathcal{L}_{\text{T5}} \;=\;
  \frac{1}{|\mathcal{L}|}\sum_{\ell \in \mathcal{L}}
  \mathcal{L}_{\text{T3}}\!\bigl(H^{(\ell)}\bigr),
  \label{eq:dst}
\end{equation}
where $H^{(\ell)}$ is the hidden-state output of layer $\ell$.
Default $\mathcal{L} = \{4, 8, 12, 16\}$ for Llama-3.2-1B
(every fourth transformer layer plus the final). The note's
formulation prescribes meta-learned per-layer weights $\lambda_\ell$;
we use uniform weights and defer meta-gradient to future work.

\subsection{T6 -- Multi-Scale Tube Bundle (MSTB-JFR)}
Generalise the central second difference to a strided stencil at
scale $\Delta$,
$
  D^2_\Delta h_t \;=\; h_{t+\Delta} - 2 h_t + h_{t-\Delta},
$
and average the \emph{$\Delta$-normalised} second difference
$D^2_\Delta h_t / \Delta^2$ across a scale set $\mathcal{S}$:
\begin{equation}
  \mathcal{L}_{\text{T6}} \;=\;
  \frac{1}{|\mathcal{S}_{\text{used}}|}\!\!
  \sum_{\Delta \in \mathcal{S}_{\text{used}}}\!\!
  \frac{1}{|\mathcal{V}_\Delta|}\!\!
  \sum_{(b,t) \in \mathcal{V}_\Delta}\!\!
  \bigl\lVert \Delta^{-2}\, D^2_\Delta h_{b,t}\bigr\rVert_2^2,
  \label{eq:mstb}
\end{equation}
where $\mathcal{V}_\Delta = \{(b,t) : \Delta \le t \le L_b - 1 - \Delta\}$
is the valid-stencil set for scale $\Delta$, and
$\mathcal{S}_{\text{used}}$ excludes any $\Delta$ for which
$\mathcal{V}_\Delta$ is empty. The
$\Delta^{-2}$ normalisation reflects the leading order of the
strided second difference for a smooth trajectory,
$D^2_\Delta h_t \approx \Delta^{2}\,\ddot h(t)$, so that all scales
contribute on a comparable scale and $\mathcal{S} = \{1\}$ recovers
T3 \eqref{eq:jfr} bit-for-bit. Default $\mathcal{S} = \{1, 2, 3\}$
(NL-RX assistant spans are typically $5\text{--}30$ tokens, so
larger scales rarely fit any example).

\subsection{T7 -- Contrastive Tube}\label{sec:methods:t7}
T7 is the only loss in our menu that injects negative pressure.
For each example, split the assistant span into two halves at
$\text{mid}_b = \text{lo}_b + \lfloor L_b / 2\rfloor$ and
mean-pool the hidden states inside each half:
\begin{equation}
  \mu^A_b = \tfrac{1}{|\text{half}_A|}\!\!\sum_{t \in \text{half}_A} h_{b,t},
  \qquad
  \mu^B_b = \tfrac{1}{|\text{half}_B|}\!\!\sum_{t \in \text{half}_B} h_{b,t}.
\end{equation}
A shared 2-layer MLP projector $g: \R^D \to \R^D \to \R^P$ with
$P = 128$ produces $z^X_b = g(\mu^X_b)$, and we $L_2$-normalise
to $\hat z^X_b = z^X_b / \norm{z^X_b}$. With temperature $\tau =
0.07$ the loss is symmetric InfoNCE with in-batch negatives:
\begin{equation}
\begin{aligned}
  \mathcal{L}_{\text{T7}}
  &= \frac{1}{2B}\sum_b \biggl[
    \mathrm{CE}\!\bigl(\hat z^A_b\,\hat Z^{B\T}/\tau,\,b\bigr) \\
  &\hspace{4em}+\;\mathrm{CE}\!\bigl(\hat z^B_b\,\hat Z^{A\T}/\tau,\,b\bigr)
  \biggr],
  \label{eq:t7}
\end{aligned}
\end{equation}
where $\hat Z^X = [\hat z^X_1,\ldots,\hat z^X_B]^\T$ and
$\mathrm{CE}$ is the cross-entropy with class label $b$ (the
diagonal). $B$ here is the \emph{effective} batch size after
dropping examples whose assistant span is too short to produce
two non-empty halves (median NL-RX span is 12 tokens, so this
dropping is rare in practice). When fewer than two examples
survive, the loss is set to zero with attached gradient.
Unlike \eqref{eq:stp}--\eqref{eq:jfr}, gradient does not vanish
when the positive pair is well-aligned: as long as $\hat z^A_b$
is closer to some other example's $\hat z^B_{b'}$ than to its
own $\hat z^B_b$, the loss is nonzero.

\subsection{T9 -- Tube-Projected Decoding (TPD)}\label{sec:methods:t9}
T9 is the single inference-time intervention we test. At
each greedy decoding step, given the previous projected hidden
state $h_{t-1}^{\,\text{proj}}$ and the current model output
$h_t^{\,\text{raw}} = f(x_{\le t})$, the projector replaces
$h_t^{\,\text{raw}}$ with
\begin{equation}
  h_t^{\,\text{proj}} \;=\; h_{t-1}^{\,\text{proj}}
  \;+\; R_\varepsilon\!\bigl(h_t^{\,\text{raw}} - h_{t-1}^{\,\text{proj}}\bigr),
  \label{eq:tpd}
\end{equation}
where the retraction $R_\varepsilon: \mathbb{R}^D \to \mathbb{R}^D$
shrinks the component of its argument orthogonal to the local
tangent (estimated from a $k$-step history of projected states)
by a factor $\alpha(\lVert v_\perp\rVert / \varepsilon)$ and
leaves the tangential component intact. The tangent direction
and tube radius $\varepsilon$ are taken from the trained
checkpoint; T9 introduces no additional trainable parameters.
The projected hidden state $h_t^{\,\text{proj}}$ is then fed to
the language-model head for the next-token argmax, while the
key/value cache used by subsequent decoding steps is built from
the un-projected forward pass (so the projector intervenes only
on the LM-head input, not on the attention state).

\paragraph{Why these nine.}
STP and T2 differ in what inner product is used for the cosine.
T1 and T3 differ in which component of the second difference is
penalised (T1 only the orthogonal-to-chord part, T3 the full
vector). T5 and T6 are depth- and scale-extensions of T3
respectively; T3-Local replaces the batch centroid with a
retrieved one and is the closest variant to the EMA-target
asymmetry of the JEPA-vision lineage. T7 differs from all
seven by being contrastive rather than reconstructive. T9
differs from all eight training-time losses by intervening only
at decoding. Together they span the attractor-vs-contrastive,
first-vs-second-order, depth, scale, retrieval-augmented-
centroid, and inference-vs-training axes that the JEPA
literature naturally suggests.

\subsection{Tier-1: Distributional Auxiliaries Outside CE's Implicit-Bias Kernel}\label{sec:methods:tier1}

Sections~\ref{sec:results:lam-sweep} and
\ref{sec:discussion:diagnosis} together motivate a sharper
question: do auxiliaries that are \emph{provably} not in
cross-entropy's implicit-bias direction --- because they
constrain the \emph{distribution} of hidden states or
tangents rather than chord-alignment --- escape the present
null? We add four such losses, all distributional rather than
geometric, and test them under the same harness as the trajectory-shape auxiliaries.
The closed-form building blocks are:

\paragraph{Epps--Pulley empirical-CF distance to $\mathcal{N}(0,1)$.}
For a 1-D sample $\{u_i\}_{i=1}^{N}$, the Plancherel-closed-form
of the squared empirical-characteristic-function distance to
the standard normal under Gaussian weight $e^{-t^2}$ is
\begin{equation}
\begin{aligned}
T_{\mathrm{EP}}(\{u_i\}) \;=\; & \frac{1}{N^2}\sum_{i,j} e^{-(u_i - u_j)^2/4} \\
& {} -\; \frac{2\sqrt{2/3}}{N}\sum_i e^{-u_i^2/6}
   \;+\; \frac{1}{\sqrt{2}}.
\end{aligned}
\label{eq:epppulley}
\end{equation}
$T_{\mathrm{EP}} \xrightarrow{\text{a.s.}} 0$ iff the sample
distribution is $\mathcal{N}(0,1)$. The two non-pair terms come
from the closed-form integrals $E_{Y \sim
\mathcal{N}(0,1)}\!\bigl[e^{-(x-Y)^2/4}\bigr] = \sqrt{2/3}\,
e^{-x^2/6}$ and $E_{Y, Y'}\!\bigl[e^{-(Y-Y')^2/4}\bigr] = 1/\sqrt{2}$.

\paragraph{Sketcher.} A linear projector
$\psi: \R^D \to \R^{d'}$ with $d' = 64$, no bias, initialised
either with a small ($\sim 10^{-2}$) Gaussian (L1, L2) or with
the Xavier scale $D^{-1/2}$ (L4). For L4 we additionally
\emph{freeze} $\psi$ (no gradient), making the projection a
random measurement map; an unfrozen L4 projector has a trivial
zero-loss fixed point at $\psi \equiv 0$ (both half-distributions
collapse to a delta at the origin).

\paragraph{L1 -- SIGReg-State.}\label{sec:methods:l1}
Stack the EOS-clipped assistant-span hidden states into the cloud
$\{h_i\}_{i=1}^{N}$, project through $\psi$, and apply the
Cram\'er--Wold mean of \eqref{eq:epppulley} over $M = 64$ random
unit directions:
\begin{equation}
  \mathcal{L}_{\text{L1}} = \frac{1}{M}\sum_{\ell}
  T_{\mathrm{EP}}\!\bigl(\{a_\ell^\T \psi(h_i)\}_i\bigr).
  \label{eq:sigregstate}
\end{equation}
By Cram\'er--Wold, $\mathcal{L}_{\text{L1}} \to 0$ iff the
projected cloud is $\mathcal{N}(0, I_{d'})$.  This is a
\emph{distributional} constraint on the marginal hidden-state
cloud, not on chord/cosine geometry of any single trajectory; by
construction it is outside CE's $L_2$-max-margin implicit-bias
direction~\cite{soudry2018}.

\paragraph{L2 -- SIGReg-Tangent.}\label{sec:methods:l2}
Apply L1 to the unit-normalised step tangents
$\Delta_{b,t} = (h_{b,t+1} - h_{b,t}) / \norm{h_{b,t+1} - h_{b,t}}$
on the assistant span:
\begin{equation}
  \mathcal{L}_{\text{L2}} = \frac{1}{M}\sum_{\ell}
  T_{\mathrm{EP}}\!\bigl(\{a_\ell^\T \psi(\Delta_i)\}_i\bigr).
  \label{eq:sigregtangent}
\end{equation}
L2 forces the tangent distribution onto the maximum-entropy
distribution on $\mathbb{S}^{D-1}$ (in projection): where T1, T3
fix the trajectory's \emph{mean} velocity, L2 fixes its
distributional \emph{shape}.

\paragraph{L3 -- C-Tube-Sectional.}\label{sec:methods:l3}
For each example sample $K = 4$ random sorted triples
$(s_k, r_k, t_k)$ inside the assistant span and define the
discrete sectional curvature
$\kappa_{\text{sec}}(h_s, h_r, h_t) = \norm{h_r - \tfrac{1}{2}(h_s+h_t)}^2 /
\tfrac{1}{4}\norm{h_t - h_s}^2$; the loss is the across-triple
variance,
\begin{equation}
  \mathcal{L}_{\text{L3}} \;=\; \mathrm{Var}_{(b,k)}\!\bigl[\kappa_{\text{sec}}\bigr].
  \label{eq:sectional}
\end{equation}
T1 \eqref{eq:ctube} drives $\kappa_{\text{sec}}$ to zero (discrete
geodesic); L3 drives its variance to zero (constant-curvature
local model, in the Killing--Hopf sense).

\paragraph{L4 -- STP-CMF.}\label{sec:methods:l4}
Split the unit-normalised tangents into two halves
$\Delta^A_b, \Delta^B_b$ at the span midpoint, project through a
\emph{fixed} sketcher $\psi_c$ (trainable $\psi_c$ has the
trivial $\psi_c \equiv 0$ escape), and compare the two halves
under the same Gaussian-kernel CF distance as~\eqref{eq:epppulley}:
\begin{equation}
  \mathcal{L}_{\text{L4}} = \frac{1}{B M}\sum_{b,\ell}
  D_{\text{CF}}\!\bigl(\{a_\ell^\T \psi_c(\Delta^A_{b,j})\},
                        \{a_\ell^\T \psi_c(\Delta^B_{b,j})\}\bigr),
  \label{eq:stpcmf}
\end{equation}
with $D_{\text{CF}}(X, Y) = \langle K\rangle_{XX} - 2\langle K\rangle_{XY} + \langle K\rangle_{YY}$
and $K(z) = e^{-z^2/4}$.  L4 is a per-example stationarity test
on the tangent distribution: STP \eqref{eq:stp} asks for matched
\emph{point} chords; L4 asks for matched \emph{distributions}.

\paragraph{Why these four.}
Each provably cannot be satisfied by motion in CE's implicit-bias
direction alone, sidestepping the orthogonality observation of
Section~\ref{sec:discussion:diagnosis}.  L1, L2 instantiate the
LeJEPA prescription at two natural target distributions (state,
tangent); L3 tests whether the geometry/distribution distinction
matters at second order; L4 lifts STP from a per-triple to a
per-trajectory distributional statement.

\subsection{Tier-2: Predictor-Based and Density-Based Auxiliaries}\label{sec:methods:tier2}

The Tier-1 question is whether \emph{any} distributional
constraint outside CE's implicit-bias direction lifts
exact-match accuracy. To make that test less specific to a
single distance choice (Cram\'er--Wold via Epps--Pulley) and
to bring in the architectural component that
Section~\ref{sec:discussion:diagnosis}-(4) names as the
missing JEPA core --- the predictor + EMA-target + stop-gradient
triple --- we add six further auxiliaries grouped into three
substructural families:

\noindent\emph{Distributional probes other than the empirical
characteristic function.} L5 (VICReg variance--covariance) and
L6 (sliced-Wasserstein isotropy) reuse L1's projector geometry
but swap the per-direction Epps--Pulley CF distance for the
two best-known anti-collapse alternatives. L9 (score matching)
uses Hyvärinen sliced score matching to estimate the score of
the projected hidden-state distribution and an additive
deviation-from-$\mathcal{N}(0, I)$ score penalty.

\noindent\emph{Predictive contrastive structure.} L12 (CPC)
asks the auxiliary head to predict $h_{t+k}$ from $h_t$ via a
small linear predictor, with negatives drawn from the same
position of every other batch example.

\noindent\emph{Predictor + target-encoder asymmetry.} L13
(BYOL-LLM) and L14 (I-JEPA-LLM) carry the predictor +
stop-gradient + EMA-target triple into our pipeline. L13 uses
mean-pooled span halves as positive pairs with an
EMA-tracked target projector; L14 masks a contiguous block of
assistant-span positions and predicts the masked targets from
a pooled summary of the visible positions.

\paragraph{L5 -- VICReg-VC.}\label{sec:methods:l5}
With $z = \psi(H_{\text{flat}})$ from the L1 sketcher, combine a
variance hinge with an off-diagonal covariance penalty:
\begin{equation}
\begin{aligned}
\mathcal{L}_{\text{L5}} =\;& \tfrac{1}{d'}\sum_{j} \max\!\bigl(0,\, 1 - \sqrt{\mathrm{Var}(z_{:,j}) + \varepsilon}\bigr) \\
& + \mu\cdot\tfrac{1}{d'(d'-1)} \sum_{j\ne k} C_{j,k}^{2},
\end{aligned}
\label{eq:vicregvc}
\end{equation}
with $C$ the empirical covariance of $z$ and $\mu = 1$ the
standard VICReg weight.  Same projector geometry as L1; the
distributional probe is direct second-moment matching rather
than a CF distance.

\paragraph{L6 -- Sliced-Wasserstein Isotropy.}\label{sec:methods:l6}
For $M$ random unit directions $a_\ell$, sort the projected
samples $u_\ell = \mathrm{sort}(\{a_\ell^\T \psi(h_i)\})$ and
compare to the standard-normal quantiles at midpoints
$q_i = \Phi^{-1}((i - 1/2)/N)$:
\begin{equation}
  \mathcal{L}_{\text{L6}} = \tfrac{1}{M}\sum_\ell \tfrac{1}{N}\sum_i (u_\ell^{(i)} - q_i)^2.
  \label{eq:swiso}
\end{equation}
This is the midpoint-quadrature estimator of
$W_2^2(F_n, \mathcal{N}(0,1))$~\cite{bonneel2015slicedwasserstein,kolouri2019gsw};
cost $O(M N \log N)$ vs.\ L1's $O(M N^2)$.  Same null as L1.

\paragraph{L9 -- Score Matching.}\label{sec:methods:l9}
Train a score net $s_\theta : \R^D \to \R^D$ via
Hyv\"arinen's identity~\cite{hyvarinen2005ssm} and add a
deviation-from-$\mathcal{N}(0, I)$ penalty:
\begin{equation}
\begin{aligned}
\mathcal{L}_{\text{L9}} =\;& \E_{z}\!\bigl[\tfrac{1}{2}\norm{s_\theta(z)}^2 + \mathrm{tr}\,\nabla_z s_\theta(z)\bigr] \\
& + \lambda_{\text{SM}}\,\E_{z}\!\bigl[\norm{s_\theta(z) + z}^2\bigr],
\end{aligned}
\label{eq:scorematch}
\end{equation}
with $\mathrm{tr}\,\nabla_z s_\theta$ estimated by Hutchinson
with one Rademacher probe per step~\cite{song2020sliced}.  The
first term recovers $\nabla_z \log p(z)$ at its minimum; the
second is the only piece that pushes the backbone toward
isotropy.

\paragraph{L12 -- Contrastive Predictive Coding.}\label{sec:methods:l12}
With $g_k : \R^D \to \R^P$ a learned linear predictor and
$t$ the midpoint of each example's valid range, train an
in-batch InfoNCE between $h_{b,t}$ and $h_{b,t+k}$ with
negatives drawn from the same position of every other batch
example:
\begin{equation}
  \mathcal{L}_{\text{L12}} = -\tfrac{1}{B'}\sum_b \log\!\frac{\exp(g_k(h_{b,t})^\T h_{b,t+k}/\tau)}{\sum_{b'}\exp(g_k(h_{b,t})^\T h_{b',t+k}/\tau)},
  \label{eq:cpc}
\end{equation}
$\tau = 0.07$.  L12 differs from T7 (Section~\ref{sec:methods:t7})
on two axes: contrast is across \emph{positions} not span
halves, and the structure is predictive not reconstructive.

\paragraph{L13 -- BYOL-LLM.}\label{sec:methods:l13}
Mean-pool each example's assistant span into halves
$\mu^A_b, \mu^B_b$.  With an online projector $f_\theta$ and
predictor $q_\phi$ and a target projector
$f_{\theta_{\mathrm{ema}}}$ updated by EMA
($\tau = 0.996$), the loss is symmetric BYOL:
\begin{equation}
\begin{aligned}
\mathcal{L}_{\text{L13}} =\;& \tfrac{1}{2B'}\sum_b\!\bigl[\norm{\widehat{q_\phi(f_\theta(\mu^A_b))} - \mathrm{sg}\,\widehat{f_{\theta_{\mathrm{ema}}}(\mu^B_b)}}^2 \\
& \quad + (A\leftrightarrow B)\bigr],
\end{aligned}
\label{eq:byol}
\end{equation}
$\widehat{x} = x/\norm{x}$.  L13 is the only auxiliary we test
containing the predictor + EMA-target + stop-gradient triple
the JEPA literature identifies as the missing JEPA core.

\paragraph{L14 -- I-JEPA-LLM (Single-Pass Simplification).}\label{sec:methods:l14}
For each example, mask a contiguous block $\mathcal{M}_b$
($\rho = 0.25$ of the assistant span), pool the visible
positions into $\bar h^{\mathrm{vis}}_b$, and train a predictor
$q_\phi$ to predict frozen Xavier-target-projected masked
states:
\begin{equation}
  \mathcal{L}_{\text{L14}} = \tfrac{1}{|\mathcal{M}|}\!\!\sum_{(b,t)\in\mathcal{M}}\!\!\norm{q_\phi(\bar h^{\mathrm{vis}}_b, \mathrm{posemb}(t)) - \mathrm{sg}\,f_\star(h_{b,t})}^2.
  \label{eq:ijepa}
\end{equation}
Faithful I-JEPA would use an EMA backbone with a second forward
pass; the single-pass approximation freezes $f_\star$ at random
init for the same anti-collapse reason as L4 (a trainable
$f_\star$ has the trivial $f_\star \equiv 0$ escape).

\paragraph{Why these six.}
The Tier-2 set tightens the parsimonious reading along the
two axes Tier-1 left underspecified: which distance one
chooses for the distributional probe (L5 vs L6 vs L1; L9 adds
a score-based probe), and whether the missing
predictor + target asymmetry is the active ingredient in the
JEPA-vision lift (L12, L13, L14). L13 in particular is the
single experiment whose result is most directly informative
about Section~\ref{sec:discussion:diagnosis}-(4): a clear lift
would imply the predictor + EMA-target asymmetry is what made
LLM-JEPA work~\cite{huang2025llmjepa}, and a null tightens
the reading further. As reported in
Section~\ref{sec:results:tier1}, the L13 cell at $n = 3$ on
TURK lands at $\Delta = +0.60$ ($p_{\text{paired}} = 0.64$); the
asymmetry mechanism therefore does not, on its own, lift exact
match in this single-pass harness.

\section{Experimental Setup}\label{sec:setup}

\paragraph{Datasets.}
We use NL-RX-TURK ($n_{\text{train}} = 8000$, $n_{\text{test}} =
2000$, evaluated on the first 500 test examples) and
NL-RX-SYNTH (same sizes), both following the chat-message
format used by~\cite{huang2026stp}: a system prompt
\textit{``Convert natural language to regular expression.''}, a
user description, and an assistant regex.

\paragraph{Model.}
Llama-3.2-1B-Instruct fine-tuned with LoRA~\cite{lora2021},
rank $r = 32$,
$\alpha = 64$, dropout $0$, applied to all attention and
multi-layer-perceptron projections (\verb|q,k,v,o,gate,up,down|).
The base model is frozen; only LoRA adapters and any auxiliary
head are trained. Mixed-precision bf16 inference and training,
with gradient checkpointing enabled. Optimiser: AdamW, peak
learning rate $2 \times 10^{-4}$, weight decay $0.01$, cosine
schedule.

\paragraph{Training schedule.}
Batch size $4$, $4$ epochs, max sequence length $256$. Each
$(\text{experiment}, \text{seed})$ cell trains in roughly
$50\text{--}60$ minutes on a single workstation; the full sweep
fits in approximately one machine-day.

\paragraph{$\lambda$ schedule.}
Each auxiliary loss is weighted by the schedule
\begin{equation}
\lambda(t) =
\begin{cases}
\lambda_0\,\dfrac{t}{T_w} & t < T_w \\
\lambda_0 & T_w \le t < T - T_d \\
\lambda_0\!\left(1 - (1{-}\rho)\dfrac{t - (T{-}T_d)}{T_d}\right) & t \ge T - T_d
\end{cases}
\end{equation}
with warmup fraction $T_w/T = 0.25$, decay fraction $T_d/T =
0.25$, and floor ratio $\rho = 0.1$ (so the final weight is
$\lambda_0 \rho$). The warmup keeps the cross-entropy
clean early so the end-of-turn classifier locks in; the decay
keeps the final logits dominated by the cross-entropy loss.
Peak weights: $\lambda_0 = 1.0$ for STP, T1, T2, T5, T7;
$\lambda_0 \in \{3 \times 10^{-4},\,10^{-3}\}$ for T3, whose
squared-second-difference output is roughly $10^3\times$
larger than the cosine-based losses (the two values come from
the $\lambda$ sweep in Section~\ref{sec:results:lam-sweep}).

\begin{figure}[t]
\centering
\includegraphics[width=\columnwidth]{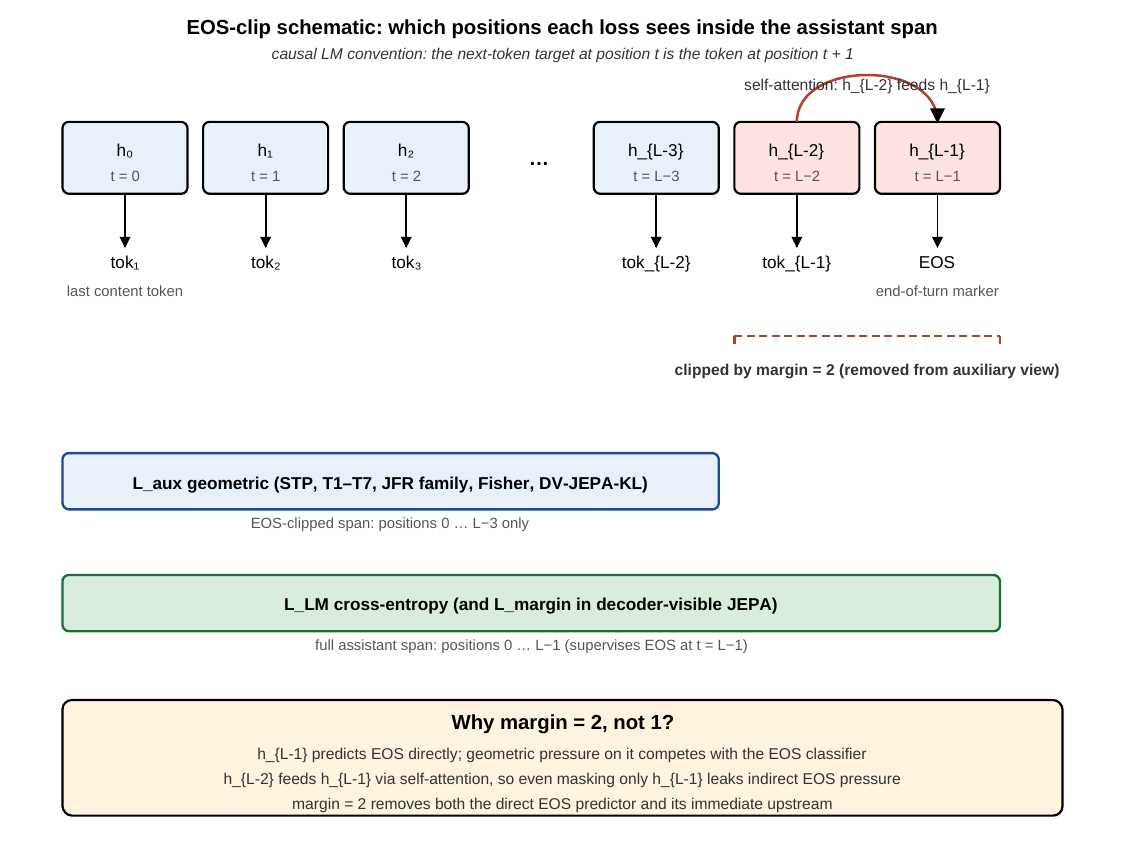}
\caption{Visibility of each loss inside the assistant span.
$h_{L-1}$ predicts EOS directly; $h_{L-2}$ feeds it via
self-attention.  Geometric auxiliaries see only the EOS-clipped
span $\{h_0, \dots, h_{L-3}\}$; cross-entropy and the
decoder-visible margin hinge see the full span and supervise EOS.}
\label{fig:eos-clip}
\end{figure}

\paragraph{Auxiliary span clip (the ``EOS clip'').}
Under causal-LM convention $h_{L_b - 1}$ predicts the end-of-turn
marker, so any geometric auxiliary acting on it competes with the
LM head's EOS classifier on the same vector.  Without
intervention this produced catastrophic EOS suppression
empirically: every auxiliary variant fell to $0\%$ exact-match
accuracy while the baseline produced $4.2$--$35.8\%$.  We
therefore clip the right end of the assistant span by
$\mathrm{margin} = 2$ tokens before passing it to any geometric
auxiliary, removing both $h_{L_b - 1}$ and the position
immediately upstream of it in attention, $h_{L_b - 2}$.
$\mathrm{margin} = 1$ was tried first and proved insufficient;
$\mathrm{margin} \geq 3$ is unnecessary on NL-RX, whose targets
do not end in a multi-token sentinel.  The clip is applied
identically to every variant under the label
\emph{EOS-clipped span}.  The token-margin hinge of
decoder-visible JEPA (Appendix~\ref{appx:dvjepa}) is the single
exception: its gradient lies inside cross-entropy's positive cone
by construction, so it cooperates with the EOS classifier and is
not clipped.  Figure~\ref{fig:eos-clip} illustrates the
construction.

\paragraph{Metrics.}
\emph{Exact-match accuracy} is strict string equality between
the greedy completion and the gold regex (modulo
\verb|skip_special_tokens=True| during decoding). \emph{Prefix
accuracy} is whether the gold regex is a prefix of the
generated completion; reporting both decouples the
``learned the regex'' signal from the ``learned to stop''
signal. Each $(\text{experiment},\,\text{seed})$ cell is evaluated
on the first $500$ test examples; we report mean and sample
standard deviation over the seeds available for that cell.

\subsection{Statistical Methodology}\label{sec:setup:stats}
For each $(\text{variant},\,\text{seed})$ cell we report two
two-sided test statistics against the matched no-auxiliary
baseline: the unpaired Welch's $t$-test
$t = (\bar x - \bar y)/\sqrt{s_x^2/n_x + s_y^2/n_y}$ with
Welch--Satterthwaite df, and the paired-by-seed Welch's
$t$-test $t = \bar d / (s_d/\sqrt{n})$ with $n - 1$ df on the
per-seed differences $d_i = x_i - y_i$.  Both are reported
because they can disagree under low $n$ when seed-level
baseline noise dominates the auxiliary effect.

We adopt $\alpha = 0.10$ as the single-cell rejection threshold
and apply Bonferroni / Holm--Bonferroni at $\alpha/k$ whenever a
family of related cells is interpreted jointly (e.g.\ the
four-cell Tier-1 SYNTH family at $\alpha/4 = 0.025$).  At
$n = 3$ the paired test has $\nu = 2$ df: the $\alpha = 0.05$
critical $t = 4.30$ requires $\sim 3.7$~pp at the observed
within-cell sd ($\sim 1.5$~pp), whereas $\alpha = 0.10$ admits
$\sim 2.5$~pp and the largest mean lifts in the study
(e.g.\ T3-Local at $+2.53$~pp) become detectable.  We treat
single-cell $\alpha = 0.10$ as a compute-gating decision for
follow-up, not as a publication-grade standard.

Cells with only one completed seed are reported under the
\emph{escalation protocol}: a single seed is used to decide whether to
expand a variant to $n = 3$, but no $t$-test, $p$-value, or
significance claim is made on $n = 1$ alone. These cells are flagged
in every table and excluded from family-level corrections.

\paragraph{A note on significance testing as a methodological choice.}
Most evaluations of representation-learning auxiliaries in the
fine-tuning literature report point estimates and seed standard
deviations without seed-level paired significance testing or
family-wise correction. Under that convention several auxiliaries in
this study would have positive face value: the largest single-seed
cells (e.g.\ $+2.73$~pp on L5 at $n=1$, the $+2.53$~pp seed-1 lift on
T3-Local) would be reported as headline gains, and seed-1 leaders
that collapse on expansion to $n = 3$ (Section~\ref{sec:results:turk-main})
would not be flagged. The conclusions in this paper are therefore a
function of the test budget --- dataset, seed count $n = 3$,
$\alpha = 0.10$ single-cell threshold, and Bonferroni / Holm at
$\alpha/k$ for joint families --- and would shift under a different
budget. Readers should treat the auxiliaries listed here as
\emph{practical tools for fine-tuning}, whose value is contingent
on dataset and compute envelope, rather than as controlled,
budget-independent results.

\section{Results}\label{sec:results}

Tables~\ref{tab:turk} and~\ref{tab:synth} report exact-match and
prefix-match accuracy on TURK and SYNTH respectively. The TURK table is
the primary result: its baseline is not saturated, and it includes the
full set of twelve training-time auxiliaries plus T9. SYNTH is used as
a secondary check because the no-auxiliary baseline is already near its
ceiling.

\begin{table*}[t]
\centering
\caption{Exact-match and prefix accuracy on NL-RX-TURK at $n = 3$ seeds,
with unpaired and paired Welch's $t$-test $p$-values against the
no-auxiliary baseline.}
\label{tab:turk}
\small
\setlength{\tabcolsep}{4pt}
\begin{tabular}{lccccc}
\toprule
Variant & Exact \% & Prefix \% & $\Delta$ & $p_\text{unp}$ & $p_\text{paired}$ \\
\midrule
Regular (no auxiliary, baseline)               & $50.67 \pm 1.68$ & $50.80 \pm 1.56$ & --       & --     & --     \\
\midrule
\multicolumn{6}{l}{\emph{Trajectory-shape auxiliaries and inference-time T9}} \\
STP                                            & $51.27 \pm 1.70$ & $51.47 \pm 1.70$ & $+0.60$  & $0.69$ & $0.37$ \\
T2 (RIG-Tube)                                  & $50.53 \pm 0.58$ & $50.73 \pm 0.58$ & $-0.13$  & $0.85$ & $0.85$ \\
T3 (JFR), $\lambda_0 = 10^{-3}$                & $52.53 \pm 0.90$ & $52.67 \pm 0.90$ & $+1.86$  & $0.19$ & $0.22$ \\
T3 (JFR), $\lambda_0 = 3{\times}10^{-4}$       & $52.93 \pm 0.64$ & $53.07 \pm 0.58$ & $+2.27$  & $0.13$ & $0.11$ \\
T3-Local (Prompt-Local JFR, $M\!=\!512$)       & $53.20 \pm 1.91$ & $53.40 \pm 1.91$ & $+2.53$  & $0.16$ & $\mathbf{0.003}$ \\
T5 (Deep Semantic Tubes, JFR base)             & $52.20 \pm 1.20$ & $52.40 \pm 1.10$ & $+1.53$  & $0.27$ & $0.08$ \\
T6 (Multi-Scale Tube Bundle, JFR base)         & $52.80 \pm 1.40$ & $53.00 \pm 1.40$ & $+2.13$  & $0.17$ & $0.16$ \\
T7 (Contrastive Tube)                          & $50.73 \pm 1.10$ & $50.87 \pm 1.01$ & $+0.07$  & $0.95$ & $0.97$ \\
\midrule
\multicolumn{6}{l}{\emph{Tier-1 distributional auxiliaries (L1--L4)}} \\
L1 (SIGReg-State)                              & $51.40 \pm 0.53$ & $51.53 \pm 0.50$ & $+0.73$  & $0.53$ & $0.47$ \\
L2 (SIGReg-Tangent)                            & $50.87 \pm 1.14$ & $51.00 \pm 1.13$ & $+0.20$  & $0.87$ & $0.91$ \\
L3 (C-Tube-Sectional)                          & $51.20 \pm 1.11$ & $51.33 \pm 1.10$ & $+0.53$  & $0.67$ & $0.77$ \\
L4 (STP-CMF)                                   & $51.07 \pm 0.83$ & $51.20 \pm 0.80$ & $+0.40$  & $0.74$ & $0.79$ \\
\midrule
\multicolumn{6}{l}{\emph{Tier-2 predictor-based and density-based auxiliaries (L5--L14)}} \\
L5 (VICReg-VC)                                 & $51.07 \pm 2.40$ & $51.20 \pm 2.40$ & $+0.40$  & $0.83$ & $0.88$ \\
L6 (SW-Iso)                                    & $51.27 \pm 0.95$ & $51.40 \pm 1.06$ & $+0.60$  & $0.63$ & $0.48$ \\
L9 (Score-Match)                               & $49.07 \pm 0.61$ & $49.27 \pm 0.61$ & $-1.60$  & $0.24$ & $0.35$ \\
L12 (CPC)                                      & $50.87 \pm 0.64$ & $51.00 \pm 0.69$ & $+0.20$  & $0.86$ & $0.77$ \\
L13 (BYOL-LLM)                                 & $51.27 \pm 0.70$ & $51.40 \pm 0.60$ & $+0.60$  & $0.61$ & $0.64$ \\
L14 (I-JEPA-LLM)                               & $49.53 \pm 1.29$ & $49.60 \pm 1.40$ & $-1.13$  & $0.41$ & $0.54$ \\
\midrule
T1 (Curvature-Aware Tube), $\lambda_0 = 1$     & $\phantom{0}2.27 \pm 0.50$ & $\phantom{0}2.47 \pm 0.50$ & $-48.40$ & --     & --     \\
T9 (TPD on T3 ckpts), $\varepsilon = 0.3$      & $\phantom{0}0.00 \pm 0.00$ & $\phantom{0}0.00 \pm 0.00$ & $-50.67$ & --     & --     \\
\bottomrule
\end{tabular}
\end{table*}

\begin{table*}[t]
\centering
\caption{Exact-match and prefix accuracy on NL-RX-SYNTH at $n = 3$ seeds,
with unpaired and paired Welch's $t$-test $p$-values against the
no-auxiliary baseline.}
\label{tab:synth}
\small
\setlength{\tabcolsep}{4pt}
\begin{tabular}{lccccc}
\toprule
Variant & Exact \% & Prefix \% & $\Delta$ & $p_\text{unp}$ & $p_\text{paired}$ \\
\midrule
Regular (no auxiliary, baseline)     & $88.93 \pm 0.42$ & $89.00 \pm 0.40$ & --      & --     & --     \\
\midrule
\multicolumn{6}{l}{\emph{Trajectory-shape auxiliaries (T1--T7)}} \\
STP                                  & $89.00 \pm 0.53$ & $89.13 \pm 0.46$ & $+0.07$ & $0.87$ & $0.83$ \\
T2 (RIG-Tube)                        & $89.47 \pm 0.83$ & $89.53 \pm 0.81$ & $+0.53$ & $0.40$ & $0.49$ \\
T3 (JFR)                             & $89.07 \pm 0.46$ & $89.13 \pm 0.58$ & $+0.13$ & $0.73$ & $0.77$ \\
T3-Local                             & $89.53 \pm 0.12$ & $89.73 \pm 0.12$ & $+0.60$ & $0.12$ & $0.19$ \\
T5 (DST-JFR)                         & $89.53 \pm 0.50$ & $89.67 \pm 0.42$ & $+0.60$ & $0.19$ & $0.19$ \\
T6 (MSTB-JFR)                        & $89.27 \pm 0.50$ & $89.47 \pm 0.50$ & $+0.33$ & $0.43$ & $0.56$ \\
T7 (Contrastive Tube)                & $88.80 \pm 0.20$ & $89.00 \pm 0.20$ & $-0.13$ & $0.65$ & $0.42$ \\
\midrule
\multicolumn{6}{l}{\emph{Tier-1 distributional auxiliaries (L1--L4)}} \\
L1 (SIGReg-State)                    & $89.80 \pm 0.53$ & $89.87 \pm 0.42$ & $+0.87$ & $\mathbf{0.09}$ & $\mathbf{0.10}$ \\
L2 (SIGReg-Tangent)                  & $89.40 \pm 0.72$ & $89.60 \pm 0.72$ & $+0.47$ & $0.40$          & $0.50$          \\
L3 (C-Tube-Sectional)                & $89.47 \pm 0.23$ & $89.67 \pm 0.23$ & $+0.53$ & $0.14$          & $\mathbf{0.09}$ \\
L4 (STP-CMF)                         & $89.73 \pm 0.70$ & $89.73 \pm 0.70$ & $+0.80$ & $0.18$          & $\mathbf{0.06}$ \\
\midrule
\multicolumn{6}{l}{\emph{Tier-2 predictor-based and density-based auxiliaries (L5--L14)$^\dagger$}} \\
L5 (VICReg-VC)                       & $89.07 \pm 0.23$ & $89.13 \pm 0.31$ & $+0.13$ & $0.66$          & $0.53$          \\
L6 (SW-Iso)                          & $89.40 \pm 0.20$ & $89.53 \pm 0.31$ & $+0.47$ & $0.18$          & $0.12$          \\
L9 (Score-Match)                     & $86.53 \pm 0.81$ & $86.60 \pm 0.80$ & $-2.40$ & $\mathbf{0.02}$ & $\mathbf{0.06}$ \\
L12 (CPC)                            & $89.20 \pm 0.53$ & $89.27 \pm 0.42$ & $+0.27$ & $0.53$          & $0.42$          \\
L13 (BYOL-LLM)                       & $88.93 \pm 0.42$ & $89.07 \pm 0.31$ & $+0.00$ & $1.00$          & $1.00$          \\
L14 (I-JEPA-LLM)                     & $88.87 \pm 0.42$ & $89.00 \pm 0.40$ & $-0.07$ & $0.85$          & $0.90$          \\
\midrule
T1 (C-Tube), $\lambda_0 = 1.0$       & $\phantom{0}9.73 \pm 0.76$ & $\phantom{0}9.80 \pm 0.87$ & $-79.20$ & $\mathbf{<0.001}$ & $\mathbf{<0.001}$ \\
\bottomrule
\end{tabular}
\end{table*}

\subsection{Main TURK Results}\label{sec:results:turk-main}

Table~\ref{tab:turk} reports the study's primary empirical content.
The no-auxiliary baseline reaches $50.67 \pm 1.68\%$ exact match.
The strongest mean training-time cell at $n = 3$ is T3-Local at
$53.20 \pm 1.91\%$ ($+2.53$~pp), with the swept single-layer T3 at
$\lambda_0 = 3 \times 10^{-4}$ next ($52.93 \pm 0.64\%$, $+2.27$~pp).
Under our pre-registered $\alpha = 0.10$ rejection threshold
(Section~\ref{sec:setup:stats}), no completed training-time
auxiliary clears the unpaired Welch test on TURK. The two paired
$t$-tests that do cross $\alpha = 0.10$ are T3-Local
($p_{\text{paired}} = 0.003$) and T5 DST-JFR ($p_{\text{paired}} = 0.08$);
neither cell's unpaired test rejects.

This is not equivalent to ``the losses do nothing''. Most
non-collapsed auxiliaries reduce seed-to-seed variance relative to
the baseline, with standard deviations falling well below the
baseline's $1.68$~pp (the L1 SIGReg-State cell tightens to $0.53$~pp,
T3 at $\lambda_0 = 3 \times 10^{-4}$ to $0.64$~pp). The most cautious
reading is that the auxiliaries regularise the hypothesis space
without reliably shifting its task-metric mean.

\subsection{Within-Family Analysis: Jacobi-Field Variants}\label{sec:results:lam-sweep}

The Jacobi-field family produces the largest mean lifts in
Table~\ref{tab:turk} and therefore admits the tightest
within-family comparison. At the matched
$\lambda_0 = 10^{-3}$ setting we observe the depth/scale ordering
\begin{equation*}
\underbrace{+1.53}_{\text{T5 (depth)}} \,<\,
\underbrace{+1.86}_{\text{T3 (single)}} \,<\,
\underbrace{+2.13}_{\text{T6 (multi-scale)}}\quad\text{pp.}
\end{equation*}
Lowering the single-layer T3 weight to $3 \times 10^{-4}$ yields
the best pure-JFR cell ($+2.27$~pp); replacing the batch
centroid with a retrieval-weighted prompt-local centroid (T3-Local)
yields the largest mean overall ($+2.53$~pp). None of these
within-family deltas crosses the unpaired $\alpha = 0.10$
threshold; the T3-Local paired test reaches
$p_{\text{paired}} = 0.003$, but with only $n = 3$ this corresponds
to $\nu = 2$ degrees of freedom and a single-cell paired result that
is not robust to the multiple-comparison correction implied by the
five JFR-family rows in the table.

Two structural conclusions follow, both qualified by $n = 3$.
\emph{(i) Depth averaging does not automatically help.} T5
underperforms matched single-layer T3 at the same $\lambda_0$,
consistent with the hypothesis that per-layer auxiliary gradients
are highly correlated under the cross-entropy backbone, so that the
depth average spends the same budget on redundant directions and
attenuates the final-layer signal.
\emph{(ii) Scale mixing and prompt-local retrieval add small but
consistently positive mean shifts.} T6 (multi-scale) and T3-Local
are the two largest means among the JFR family, in line with the
intuition that scale-invariance and EMA/prompt-local target
asymmetry inject information the single-scale, batch-centroid T3
loss does not. The most informative next step is not a sixth JFR
variant but increased seed counts and data-efficiency
curves that can adjudicate whether these small mean shifts are
stable.

\subsection{Contrastive and Distributional Auxiliaries}\label{sec:results:tier1}

The contrastive variant T7 was meant to test whether negative pressure
breaks out of the attractor-loss regime. It does not: T7 lands at
$+0.07$~pp on TURK with $p\approx0.95$. At batch size $4$, this is not
a decisive statement about contrastive learning in general, but it is a
decisive statement about this small-batch LoRA harness.

The distributional auxiliaries L1--L4 were designed as a stronger
escape test. Unlike STP/T1/T3-style trajectory losses, they constrain
marginal or tangent distributions and were explicitly constructed to
fall outside cross-entropy's implicit-bias direction. They also sit on
baseline: $\Delta_{\text{exact}}=+0.20$ to $+0.73$~pp, with every
paired and unpaired test at $p\ge0.47$. This result is important
because it rules out the simple explanation that the original geometric
losses failed only because they were too aligned with cross-entropy.
Even the out-of-kernel distributional losses do not move exact match.

\paragraph{Tier-2 results: the JEPA-core test and the escalation protocol.}
All six Tier-2 cells are now at $n = 3$ on TURK, completing the
training-time grid at three seeds. Two findings are worth flagging
explicitly. \emph{First}, the largest single-seed lift,
L5 VICReg-VC at $+2.73$~pp on seed 0, \emph{collapses} on expansion
to an $n = 3$ mean of $51.07 \pm 2.40\%$
($\Delta = +0.40$, $p_{\text{paired}} = 0.88$). The seed-0 mark was
a single-sample artefact. We treat this as the clearest
demonstration of why pre-registered escalation matters in small-$n$
experiments: a static single-seed table would have reported
$\Delta_{\text{exact}} = +2.73$ as the apparent leader, and the
protocol would have been wrong. \emph{Second}, L13 BYOL-LLM ---
the only cell we test containing the predictor + EMA-target +
stop-gradient triple that
Section~\ref{sec:discussion:diagnosis}-(4) names as the JEPA-core
ingredient missing from STP and L1--L4 --- lands at
$51.27 \pm 0.70\%$ on TURK ($\Delta = +0.60$,
$p_{\text{paired}} = 0.64$) and $88.93 \pm 0.42\%$ on SYNTH
($\Delta = 0.00$, $p_{\text{paired}} = 1.00$). The predictor +
target asymmetry that lifted the saturated SYNTH baseline by
$\sim\!6$~pp in LLM-JEPA~\cite{huang2025llmjepa} does \emph{not}
produce a detectable lift on either benchmark in our single-pass
harness. L14 I-JEPA-LLM, the second predictor + frozen-target cell,
also sits inside seed noise on both benchmarks
($\Delta_{\text{TURK}} = -1.13$, $\Delta_{\text{SYNTH}} = -0.07$).
The remaining three Tier-2 cells span $\Delta \in [-1.60, +0.60]$
on TURK with every paired and unpaired test at $p \ge 0.24$:
L6 SW-Iso ($+0.60$, $p_{\text{paired}} = 0.48$), L12 CPC
($+0.20$, $p_{\text{paired}} = 0.77$), L9 Score-Match
($-1.60$, $p_{\text{paired}} = 0.35$). No completed Tier-2 cell on
either benchmark clears $\alpha = 0.10$.

\subsection{Failure Modes: T1 Collapse and T9 Decoding}\label{sec:results:failures}

T1 at $\lambda_0=1.0$ collapses training-time performance on both
benchmarks: $2.27\pm0.50\%$ exact match on TURK and
$9.73\pm0.76\%$ on SYNTH. The collapse is therefore not a dataset
quirk. It is an interaction failure between the curvature penalty and
the language-model fine-tuning objective at that weight.

T9 is worse as an inference intervention. Activating tube-projected
decoding on the three T3 checkpoints yields $0.00\%$ exact and prefix
accuracy. Generated strings collapse into repetitive token loops, which
suggests that the retraction creates an argmax fixed point under greedy
decoding. T9 is therefore a useful negative control: making the hidden
state more tube-consistent at inference time can destroy generation.

\subsection{SYNTH Results and Family-Level Analysis}\label{sec:results:synth-cluster}

On NL-RX-SYNTH, the no-auxiliary baseline already reaches
$88.93 \pm 0.42\%$ exact match. Every completed non-collapsed
auxiliary falls within $+0.87/-0.13$~pp of that baseline; SYNTH
therefore cannot adjudicate small auxiliary effects through point
estimates alone. What it does add, once all four Tier-1
distributional auxiliaries land at $n = 3$, is a coherent
\emph{family-level} pattern: every L1--L4 cell sits above
baseline, and three of the four clear the conventional
$\alpha = 0.10$ threshold under the paired Welch's $t$-test
(L4 STP-CMF at $\mathbf{p_{\text{paired}} = 0.06}$, L3
C-Tube-Sectional at $0.09$, L1 SIGReg-State at $0.10$; only L2
SIGReg-Tangent at $0.50$ stays clearly inside noise). The
unpaired versions land at $p \in [0.09, 0.40]$. The strongest
mean lift is L1 at $+0.87$~pp, with L4 next at $+0.80$.

We do not interpret the L1--L4 cluster as a positive finding.
Three considerations argue against doing so. \emph{First}, the
absolute effect sizes ($\le 1$~pp on a benchmark whose baseline is
within $\sim\!11$ points of the ceiling) sit well below the
resolution at which exact-match is a reliable signal in this
regime. \emph{Second}, the four Tier-1 cells share the same three
baseline seeds, so the paired tests are not statistically
independent; the Bonferroni correction at the family-wise error
rate $\alpha_{\text{FWE}} = 0.10$ reduces the per-test threshold
to $0.025$, against which all four cells fail. The Holm--Bonferroni
step-down procedure on the same four ordered $p$-values (smallest
first) requires $0.057 < 0.025$ at the first step and likewise
rejects every cell. \emph{Third}, expanding to the full ten-cell distributional family
on SYNTH (L1--L6, L9, L12, L13, L14, all at $n = 3$) leaves the
qualitative picture intact: three Tier-1 cells clear paired
$\alpha = 0.10$ uncorrected, every Tier-2 distributional cell sits
comfortably inside seed noise, and one Tier-2 cell (L9 Score-Match)
clears paired $\alpha = 0.10$ \emph{with a negative delta}
($\Delta = -2.40$, $p_{\text{paired}} = 0.057$). The only cell on
SYNTH that crosses the uncorrected threshold under multiple-cell
coverage therefore points the wrong way. Bonferroni at
$0.10/10 = 0.01$ rejects all ten cells; Holm--Bonferroni on the
smallest paired $p$-value $0.057 \not< 0.01$ rejects every cell.
The best-supported conclusion is that SYNTH shows a small Tier-1
family-coherent positive drift, an L9 cell that significantly
\emph{reduces} exact match under uncorrected paired testing, and
otherwise no movement that survives correction. ``The auxiliaries
do real geometric work'' is consistent with the data; ``the work
moves exact match by a decision-grade margin'' is not. T1 collapses
on SYNTH at the same severity as on TURK ($9.73 \pm 0.76\%$,
$p < 10^{-3}$), confirming that the collapse is dataset-independent.

\subsection{Data-Efficiency Curve}\label{sec:results:dataeff}

Section~\ref{sec:discussion:future} explicitly named a
data-efficiency lift on T3-Local as one of the conditions
that would update the structured-null reading: the original STP
claim is framed as a sample-efficiency gain rather than an
asymptotic-exact-match gain, and a small-data lift would be
consistent with the auxiliaries doing real representational work
that the saturated end of the curve does not reveal. We therefore
re-train baseline, T3-Local, T6 MSTB-JFR and L4 STP-CMF on TURK
at five training-data fractions (5\%, 10\%, 25\%, 50\%, 100\% of
$n_{\mathrm{train}} = 8000$), three seeds per cell, with all other
harness conditions held fixed.

\begin{figure}[t]
\centering
\includegraphics[width=\columnwidth]{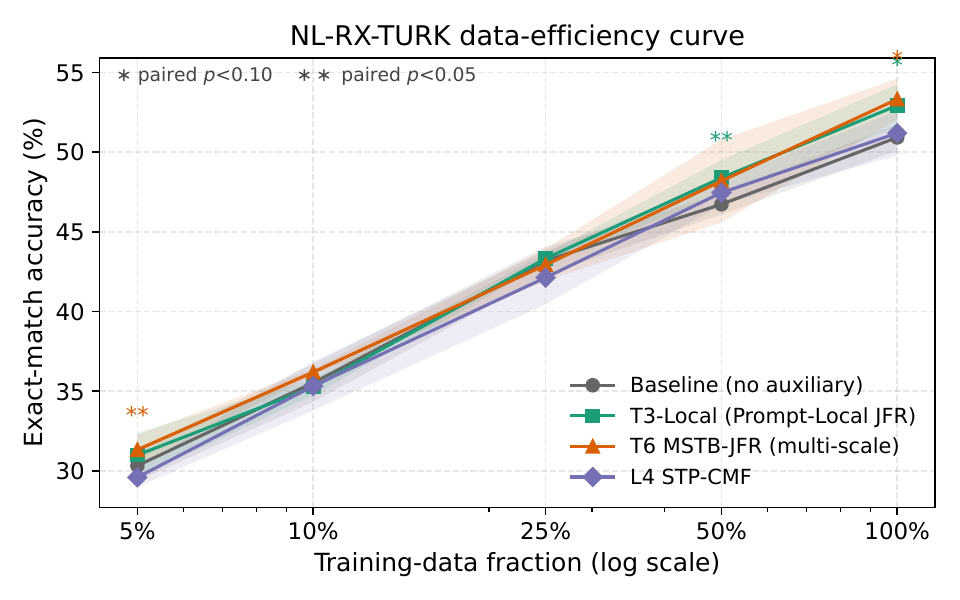}
\caption{Data-efficiency curve on NL-RX-TURK. Mean $\pm$ one
standard deviation across three seeds at each training-data
fraction; significance markers are paired Welch's $t$-tests of
each auxiliary against the matched baseline cell ($\ast$:
$p_{\mathrm{paired}} < 0.10$, $\ast\ast$:
$p_{\mathrm{paired}} < 0.05$).}
\label{fig:dataeff}
\end{figure}

Table~\ref{tab:dataeff} reports per-fraction means and paired
$t$-tests. The data resolve the paper's main question with a
\emph{regime-conditional yes}. Two distinct cells clear paired
$\alpha = 0.05$, at different fractions and from different
auxiliary families; two further cells sit just inside paired
$\alpha = 0.10$ at $100\%$ data:

\emph{(i) T6 MSTB-JFR is positive at both ends of the curve.}
At $5\%$ data ($400$ examples), T6 clears
$p_{\mathrm{paired}} = 0.013$ with $\Delta = +1.00$~pp; the lift
fades at $10\%$ ($\Delta = +0.67$, $p = 0.50$) and reverses at
$25\%$ ($\Delta = -0.27$, $p = 0.46$); but at $100\%$ data
($8000$ examples) T6 returns the largest mean lift in the cube
($\Delta = +2.40$~pp, $p_{\mathrm{paired}} = 0.059$).
The multi-scale strided JFR is positive both when training data
is scarce and at full-data saturation, but null in between.

\emph{(ii) T3-Local at the mid-data point and at saturation.}
At $50\%$ data ($4000$ examples), T3-Local clears
$p_{\mathrm{paired}} = 0.029$ with $\Delta = +1.67$~pp; the lift
persists at $100\%$ ($\Delta = +2.00$, $p = 0.090$, recovering
the effect size of the original Table~\ref{tab:turk}
measurement) but is absent in the small-data regime
($\Delta = +0.67$, $p = 0.13$ at $5\%$;
$\Delta = -0.27$, $p = 0.79$ at $10\%$).

\emph{(iii) L4 STP-CMF shows no positive effect at any of the
five fractions} ($\Delta \in [-1.07, +0.73]$, every
$p_{\mathrm{paired}} \ge 0.37$). The completed $50\%$
($\Delta = +0.73$, $p = 0.37$) and $100\%$
($\Delta = +0.27$, $p = 0.83$) cells confirm the small-data
trend: L4's distributional pressure does not, on its own,
translate into exact-match accuracy at any sample size we tested.

The two paired-$\alpha = 0.05$ cells clear on \emph{disjoint}
data fractions and come from \emph{different auxiliary families}:
T6 (multi-scale strided JFR) at $5\%$, T3-Local
(prompt-neighbour JFR) at $50\%$. This is inconsistent with a
single ``small-data sample-efficiency lift'' interpretation and
consistent with a regime-specific picture in which different
auxiliaries help in different regimes.

\begin{table*}[t]
\centering
\caption{Data-efficiency curve on NL-RX-TURK. Exact-match mean $\pm$
sample standard deviation at each fraction (three seeds); paired
Welch's $t$-test against the matched-fraction baseline.
\textbf{Bold}: paired $p < 0.05$.}
\label{tab:dataeff}
\small
\setlength{\tabcolsep}{4pt}
\begin{tabular}{rcccccccccc}
\toprule
& \multicolumn{2}{c}{Baseline}
& \multicolumn{2}{c}{T3-Local}
& \multicolumn{2}{c}{T6 MSTB-JFR}
& \multicolumn{2}{c}{L4 STP-CMF} \\
\cmidrule(lr){2-3}\cmidrule(lr){4-5}\cmidrule(lr){6-7}\cmidrule(lr){8-9}
Frac. & mean $\%$ & --
& mean $\%$ & $p_{\mathrm{paired}}$
& mean $\%$ & $p_{\mathrm{paired}}$
& mean $\%$ & $p_{\mathrm{paired}}$ \\
\midrule
$5\%$   & $30.33 \pm 0.95$ & --
        & $31.00 \pm 1.40$ & $0.13$
        & $31.33 \pm 0.92$ & $\mathbf{0.013}$
        & $29.60 \pm 0.60$ & $0.41$ \\
$10\%$  & $35.53 \pm 1.21$ & --
        & $35.27 \pm 0.50$ & $0.79$
        & $36.20 \pm 0.20$ & $0.50$
        & $35.33 \pm 1.53$ & $0.91$ \\
$25\%$  & $43.20 \pm 0.87$ & --
        & $43.33 \pm 0.31$ & $0.84$
        & $42.93 \pm 1.01$ & $0.46$
        & $42.13 \pm 1.67$ & $0.54$ \\
$50\%$  & $46.73 \pm 0.64$ & --
        & $48.40 \pm 1.11$ & $\mathbf{0.029}$
        & $48.20 \pm 2.60$ & $0.32$
        & $47.47 \pm 0.76$ & $0.37$ \\
$100\%$ & $50.93 \pm 0.83$ & --
        & $52.93 \pm 1.33$ & $0.090$
        & $53.33 \pm 1.29$ & $0.059$
        & $51.20 \pm 1.40$ & $0.83$ \\
\bottomrule
\end{tabular}
\end{table*}

These two paired-$\alpha = 0.05$ cells are this study's first
positive results on a fixed exact-match metric, but neither
clears Bonferroni or Holm--Bonferroni at $\alpha/15 \approx
0.0033$ over the cube; the cells live at \emph{different
fractions} and involve \emph{different auxiliaries}, so they are
not mutually corroborating; and L4 STP-CMF, the third
pre-registered variant, shows no positive effect at any
fraction ($\Delta \in [-1.07, +0.73]$, every $p \ge 0.37$),
breaking the family-coherence reading.  The conservative
adoption: representational work \emph{can} translate into
exact-match accuracy under specific conditions, but no cell
survives multiple-comparison correction.  The four
$p < 0.10$ cells are candidate positives to confirm with
$n \ge 5$ seeds, by adding SYNTH to the cube, and by reproducing
on a second base model.

\subsection{Summary of Empirical Findings}\label{sec:results:summary}

Six observations summarise the empirical content.
\emph{(i)} No training-time auxiliary clears the unpaired Welch
test on TURK at $\alpha = 0.10$; the strongest mean lift is
$+2.53$~pp (T3-Local), inside seed noise.
\emph{(ii)} On SYNTH the four Tier-1 distributional cells form a
small family-coherent cluster (every $\Delta \ge 0$, three of
four with paired $p < 0.10$) that does not survive Bonferroni
$\alpha/4$ or Holm--Bonferroni step-down.
\emph{(iii)} BYOL-LLM, the only cell carrying the
predictor + EMA-target + stop-gradient triple, lands inside seed
noise on both benchmarks ($\Delta_{\mathrm{TURK}} = +0.60$,
$\Delta_{\mathrm{SYNTH}} = 0.00$); the architectural asymmetry
that lifts vision JEPA does not on its own move exact match in a
single-pass harness.
\emph{(iv)} The TURK data-efficiency cube
(Section~\ref{sec:results:dataeff}) reveals two paired-$\alpha = 0.05$
cells on disjoint fractions and from different auxiliary families
(T6 at $5\%$, T3-Local at $50\%$) and two further cells just inside
paired $\alpha = 0.10$ (T6 and T3-Local at $100\%$); L4 shows no
positive effect at any fraction.  No cell survives Bonferroni at
$\alpha/15$, but four cells with paired $p < 0.10$ are this study's
first concrete instance of representational work translating into
exact-match accuracy.
\emph{(v)} The two pre-registered falsification routes both fail
to break the null on TURK: Fisher cells lose every head-to-head
against their Euclidean twins by $1.20$--$2.47$~pp and erase the
geometric signature observed in the Euclidean cells;
decoder-visible JEPA produces the study's first positive gradient
cosine but undershoots its pre-registered threshold and leaves
exact match inside seed noise.
\emph{(vi)} T1 at $\lambda_0 = 1$ and T9 collapse on both
benchmarks; L9 Score-Match \emph{reduces} SYNTH exact match by
$-2.40$~pp ($p_{\text{paired}} = 0.057$), the only Tier-2 cell
that crosses the uncorrected paired threshold.

Within the harness studied here, adding richer single-forward-pass
JEPA-style geometric, distributional, decoder-aligned, or
positive-cone regularisation is therefore not a reliable route to
improved exact-match accuracy.  The next informative directions
(Section~\ref{sec:discussion:future}) are seed expansion to
$n \geq 10$ on the candidate cells, two-view predictor
architectures that go beyond a single causal forward pass,
larger contrastive batches, and metrics that can resolve
representation-level gains directly.

\section{Discussion}\label{sec:discussion}

The results leave two questions. First, did the auxiliaries fail because
they were absorbed by cross-entropy or because they acted in directions
that exact match does not reward? Second, which existing JEPA-theory
explanation survives contact with the diagnostics?

\subsection{Diagnostic Findings}\label{sec:discussion:diagnosis}

Appendix~\ref{appx:diagnostics} reports three measurements on trained
checkpoints: assistant-span anisotropy, trajectory curvature, and the
gradient cosine between auxiliary loss and language-model
cross-entropy over LoRA-trainable parameters. Together they establish
that the main null is \emph{not} a no-op null.

First, JFR-family auxiliaries reduce trajectory curvature from the
baseline value of roughly $2.00$~rad to the $1.21$--$1.45$~rad range.
Second, several auxiliaries tighten seed-to-seed variance. Third, and
most importantly for the title, the measured gradient cosine is near
zero or slightly negative rather than near one: across the JFR-family
TURK cells, $\rho\in[-0.13,+0.01]$. Thus the auxiliary gradients are
not simply parallel to cross-entropy's update direction. They are doing
representational work that cross-entropy is not already doing.

We summarise this as an empirical diagnostic criterion.  An auxiliary is
\emph{representationally active but task-metric inert} when
\begin{equation}
  \Delta G > 0,\qquad
  \left|\rho_{\text{aux,CE}}\right| \ll 1,\qquad
  \left|\Delta \mathrm{EM}\right| \leq \epsilon,
  \label{eq:active-inert}
\end{equation}
where $\Delta G$ is a measured geometry change, such as reduced curvature
or changed anisotropy, and
\begin{equation}
  \rho_{\text{aux,CE}} =
  \frac{\inner{\nabla_{\theta_{\text{LoRA}}}\mathcal{L}_{\text{aux}}}
  {\nabla_{\theta_{\text{LoRA}}}\mathcal{L}_{\text{LM}}}}
  {\norm{\nabla_{\theta_{\text{LoRA}}}\mathcal{L}_{\text{aux}}}
   \norm{\nabla_{\theta_{\text{LoRA}}}\mathcal{L}_{\text{LM}}}}.
  \label{eq:grad-cosine}
\end{equation}
The measured cells satisfy this pattern: they change geometry,
they do not move in the CE direction, and they do not reliably
move exact match.  The diagnostics therefore weaken three simple
explanations of why exact match does not move ---
\emph{CE absorption} (the gradient-cosine measurement rules this
out), \emph{already-straight trajectories} (the curvature
diagnostic rules this out), and \emph{missing anti-collapse only}
(L1--L4 would have been the escape route, and were not).  The
remaining explanation is more specific: the losses change
geometry, but the exact-match regex metric is insensitive to the
changed geometry.  The auxiliaries act, but their action is
decoder-invisible.

\subsection{Failure-mode mechanisms}\label{sec:discussion:failures}

Two specific failures admit short mechanistic readings.  T5
averages the same JFR pressure across four transformer layers; if
the per-layer gradients were independent this could increase
signal, but the observed ordering suggests the opposite --- the
gradients are correlated under the CE backbone, so the depth
average spends the same budget on redundant directions and
dilutes the final-layer signal.  T9 composes the autoregressive
forward step with a retraction onto the learned tube; greedy
decoding is already prone to local attractors, and a non-expansive
projection strengthens that contraction into a stable argmax
fixed point, producing the repetitive-token failure observed at
$0.00\%$ exact and prefix accuracy.  A viable projector would
need to preserve an entropy-bearing direction or inject
stochasticity.

\subsection{Relation to the Original STP Claim}

Our experiments do not directly falsify STP's data-efficiency
claim~\cite{huang2026stp}: that claim is about a training-data
fraction curve on NL-RX-SYNTH, ours is about asymptotic
exact-match accuracy.  A loss can plausibly improve sample
efficiency without moving the final exact-match ceiling.
Section~\ref{sec:results:dataeff} reports the TURK arm and finds
two paired-$\alpha = 0.05$ cells on disjoint fractions from
different auxiliary families --- partial corroboration; the SYNTH
arm of that curve remains the empirical gap.  A second difference
is scale: T7 uses in-batch negatives but our setup fixes the
batch size at $4$, so the T7 null is a small-batch result, not a
general statement about contrastive LLM auxiliaries.

\subsection{The Structured-Null Interpretation}\label{sec:discussion:parsimonious}

The most compact interpretation is:
\begin{quote}
The auxiliaries do representational work without decoder-visible gain.
They reshape hidden-state geometry in directions that are not
cross-entropy's update direction, but those directions do not reliably
affect exact-match regex accuracy in this fine-tuning regime.
\end{quote}

This reading explains the main observations at once: small and
non-significant mean gains, reduced variance, curvature reduction,
gradient orthogonality, distributional-loss nulls, and the contrast
between training-time regularisation and T9's inference-time fixed-point
collapse. It also makes a clear prediction: metrics or regimes that
reward representation geometry---data efficiency, calibration,
out-of-distribution generalisation, semantic equivalence, or larger
contrastive batches---are more likely to reveal value than another
single-forward-pass trajectory penalty evaluated only by exact match.

The reading is also corroborated by an independent positive result.
Wang et al.~\cite{wang2026temporalstraightening} train a world-model
encoder with a curvature regulariser that, in our notation, is
structurally identical to STP, and report $20$--$60\%$ open-loop and
$20$--$30\%$ MPC goal-reaching gains relative to a non-regularised
baseline. They
additionally prove that, under a linear-dynamics assumption, the same
regulariser controls the condition number of the planning Hessian.
Their gain is precisely the kind that our structured-null reading
predicts \emph{should} appear: their downstream metric is the
optimisation cost of a gradient-based planner whose Hessian explicitly
depends on representation conditioning. Our downstream metric is
exact-match decoded under argmax, which has no such dependence. The
two results together suggest that whether trajectory straightening
helps a downstream task is governed less by whether the representation
is straighter and more by whether the downstream objective's
optimisation landscape is shaped by representation geometry.

We probe this reading directly with two pre-registered
falsification routes.  Appendix~\ref{appx:fisher} replaces the
Euclidean curvature norm with the LM head's pulled-back
next-token Fisher metric, additionally weights each per-token
contribution by a margin-driven sigmoid, and projects the
auxiliary gradient off any negative-conflict component of CE.
If the null mistook ``wrong direction in input space'' for ``no
signal at the decoder,'' any of these three interventions should
have recovered some exact-match accuracy.  The TURK cube
(Table~\ref{tab:fisher}) instead returns five head-to-head
regressions of $1.20$--$2.47$~pp against the Euclidean twins; on
T3-Local the regression deepens monotonically as components are
added.  Variance contracts in every Fisher cell by $14$--$68\%$,
matching the Euclidean cells: representation work is happening,
the metric just does not expose it.  Appendix~\ref{appx:dvjepa}
moves the auxiliary out of $h$-space entirely, scoring it on the
post-softmax distribution of the LM head; by construction its
margin-side gradient lies in cross-entropy's convex cone.  We
pre-registered $\rho_{\mathrm{aux,CE}} > 0.20$ on TURK at $n = 3$
as the diagnostic-level falsification threshold.  The result
(Table~\ref{tab:dvjepa}): $\rho_{\mathrm{aux,CE}} = +0.186$, four
standard deviations above the upper end of the existing band but
undershooting the pre-registered threshold; exact match remains
inside seed noise and the auxiliary leaves anisotropy and
curvature at baseline.  The directional prediction is correct,
the magnitude prediction is overconfident, and the decoded-task
null persists --- the structured-null reading is robust against
both falsification routes.

\subsection{Future Directions and Decision Criteria}\label{sec:discussion:future}

The most informative next experiments are ordered by how directly they test
the structured-null interpretation:
\begin{enumerate}
\item \textbf{Data-efficiency curves.}  We have completed the
  TURK arm at $4 \times 5 \times 3$
  (Section~\ref{sec:results:dataeff}); the SYNTH arm of the same
  five-fraction sweep is the remaining benchmark gap. Within the
  TURK cube, expanding the four cells with paired $p < 0.10$ to
  $n \ge 5$ seeds is the most informative seed-expansion
  follow-up.
\item \textbf{Seed expansion for plausible cells.}  Expand T3-Local,
  T3 at $3\times10^{-4}$, T6, L1 and L4 to $n\ge10$ before proposing new
  tube variants.
\item \textbf{Metric expansion.}  Add regex semantic-equivalence tests,
  calibration, and OOD splits to determine whether the geometry change is
  useful but invisible to exact string match.
\item \textbf{Scale the discriminative tests.}  Revisit T7 and CPC only
  with substantially larger effective batches or memory-bank negatives.
\item \textbf{Test full multi-view JEPA.}  Replace the single-pass
  approximations with a genuine context-target architecture with a target
  encoder and non-trivial view construction.
\item \textbf{Replication under full fine-tuning.} The closest prior
  comparators run their main experiments under \emph{full} fine-tuning:
  STP's $16\times$ data-efficiency claim and LLM-JEPA's headline
  accuracy gains are full-parameter results, with LoRA reported only as
  a secondary appendix study (LLM-JEPA, Table 8). The main grid in
  this paper is LoRA throughout, raising the question of whether the
  structured null is a property of the auxiliaries or of the regime.
  We tested decoder-visible JEPA, the most theoretically promising of
  the falsification routes, under full fine-tuning at $n = 5$ seeds on
  TURK with the same auxiliary hyperparameters, dropping the learning
  rate from $2{\times}10^{-4}$ (LoRA) to $2{\times}10^{-5}$ to match
  STP and LLM-JEPA's full-FT recipe. The result reproduces the LoRA
  null on both benchmarks. On TURK the no-auxiliary baseline and
  DV-JEPA produced $48.72\% \pm 0.61$~pp and $48.76\% \pm 1.42$~pp,
  with $\Delta = +0.04$~pp, $t_{\text{paired}} = +0.054$ ($\nu = 4$),
  $p_{\text{paired}} = 0.96$, and the unpaired test agreeing
  ($t = +0.058$, $p = 0.96$). On SYNTH the same comparison yielded
  $88.40\% \pm 0.47$~pp (regular) versus $88.92\% \pm 0.73$~pp
  (DV-JEPA), with $\Delta = +0.52$~pp, $t_{\text{paired}} = +1.260$
  ($\nu = 4$), $p_{\text{paired}} = 0.28$, and the unpaired test in
  agreement ($t = +1.341$, $p = 0.22$). Bonferroni correction at
  $\alpha/2 = 0.05$ across the two head-to-heads is therefore not
  approached, and neither cell clears the single-cell $\alpha = 0.10$
  threshold either. The structured null is therefore robust across
  the LoRA and full-FT regimes \emph{and} across both benchmarks for
  the decoder-visible construction, weakening the capacity-ceiling
  reading and strengthening the broader conclusion that
  hidden-geometry representation work and exact-match decoded
  accuracy are weakly coupled in autoregressive next-token training
  irrespective of which parameters carry gradient. STP-style
  straightening, which the comparators report as full-FT-positive,
  remains an open replication target; we did not run the
  trajectory-shape arm under full FT in this paper.
\end{enumerate}
The conclusion would change if any one of these settings produced a stable,
multiple-comparison-corrected gain while preserving the diagnostic evidence
that the auxiliary acts outside the cross-entropy direction.

\subsection{Limitations}

$n = 3$ seeds is small for detecting effects $\le 2$~pp at the
variances observed here, so the present study is best read as a
hypothesis test and prioritisation device, not a final power
analysis.  Results are restricted to Llama-3.2-1B-Instruct with
LoRA on the NL-RX family; larger base models, full fine-tuning,
non-symbolic tasks, or tasks with more semantic slack may behave
differently.  T3 receives a two-point $\lambda_0$ sweep, but most
auxiliaries are run at one weight, and the T1 collapse at
$\lambda_0 = 1.0$ shows that weight matters --- smaller T1
weights and broader sweeps remain useful checks.  The bf16 setup
fixes batch size at $4$ and rules out the large negative banks
contrastive objectives typically rely on, which is the dominant
limitation for interpreting T7.

\section{Conclusion}

This paper asked when JEPA-style auxiliary objectives added to LoRA
fine-tuning create decoder-visible task signal, and when they merely
reshape hidden-state geometry in directions that exact-match decoding
does not reward. We studied twenty-two training-time auxiliaries and
one inference-time projector for Llama-3.2-1B-Instruct under fixed
model, splice point, decoding, and evaluation conditions, and we
constructed two pre-registered falsification routes for the
structured-null reading itself: a Fisher-metric Jacobi-residual family
(decoder-aligned curvature) and decoder-visible JEPA (auxiliary
gradient engineered to share its convex cone with cross-entropy).

The main result is a structured null robust against both falsification
routes. Without statistical correction, a handful of cells reach
single-cell paired $\alpha = 0.10$: prompt-local JFR on TURK
($\Delta = +2.53$~pp, $p_{\text{paired}} = 0.003$), T5 on TURK,
three Tier-1 distributional cells on SYNTH (L1, L3, L4), and four
fraction-specific cells in the data-efficiency cube (T6 at $5\%$
and $100\%$, T3-Local at $50\%$ and $100\%$). With Bonferroni or
Holm--Bonferroni at the family-wise threshold, none of these
survives. BYOL-LLM, the direct test of JEPA-core predictor +
EMA-target + stop-gradient asymmetry, remains inside seed noise on
both benchmarks even without correction.  The Fisher-metric replacement loses every
head-to-head against its Euclidean twin ($-1.20$ to $-2.47$~pp on
TURK) and \emph{erases} the study's geometric signature: anisotropy
and trajectory curvature snap back to baseline, indicating that
confining the auxiliary to decoder-perceptible directions removes its
h-space work without buying back exact-match accuracy.  Decoder-visible
JEPA produces the study's first cell with positive gradient cosine,
$\rho_{\mathrm{aux,CE}} = +0.186 \pm 0.050$, four standard deviations
above the upper end of the $[-0.69, +0.01]$ band every other cell
occupies; the pre-registered $\rho > 0.20$ threshold is just missed
in the mean and exact match remains inside seed noise
($\Delta = +0.93$~pp, $p_{\mathrm{paired}} = 0.48$).  The directional
prediction is correct, the magnitude prediction is overconfident, and
the decoded-task null persists.

The null is not a representational null. The diagnostics show that
JFR-family auxiliaries reduce trajectory curvature, that L1, L6, and
L14 collapse anisotropy below baseline, and that several auxiliaries
tighten seed-to-seed variance.  In this setting, JEPA-style
auxiliaries can therefore perform measurable representational work
without producing reliable decoder-visible gains.  Moving hidden
geometry is not the same as improving the decoded task metric, and
even an auxiliary explicitly engineered to share cross-entropy's
gradient cone does not, at this seed count, translate into exact-match
accuracy.

A full-fine-tuning replication of decoder-visible JEPA on both
benchmarks at $n = 5$ seeds (Section~\ref{sec:discussion:future})
reproduces this null. With the LoRA adapters dropped and the LM
body trained end-to-end at $\text{lr} = 2{\times}10^{-5}$, the
no-auxiliary baseline and DV-JEPA reach $48.72\% \pm 0.61$~pp and
$48.76\% \pm 1.42$~pp respectively on TURK ($\Delta = +0.04$~pp,
$t_{\text{paired}} = +0.054$, $\nu = 4$, $p_{\text{paired}} = 0.96$),
and $88.40\% \pm 0.47$~pp and $88.92\% \pm 0.73$~pp on SYNTH
($\Delta = +0.52$~pp, $t_{\text{paired}} = +1.260$, $\nu = 4$,
$p_{\text{paired}} = 0.28$). Neither cell clears single-cell
$\alpha = 0.10$, much less the Bonferroni gate at $\alpha/2 = 0.05$.
The structured null is therefore robust across LoRA and full
fine-tuning, and across both benchmarks, for the decoder-visible
construction. This weakens the capacity-ceiling reading and supports
the broader claim that representation work and exact-match accuracy
are weakly coupled irrespective of which parameters carry gradient.

This conclusion narrows the design space for future LLM-domain JEPA
work along two axes.  First, the seed budget: the full cube at
$n \geq 10$ on the four candidate cells with paired $p < 0.10$
(T6@$5\%$, T6@$100\%$, T3-Local@$50\%$, T3-Local@$100\%$) and on
decoder-visible JEPA is the cheapest informative
follow-up.  Second, the metric: the structured-null reading
predicts that exact match is insensitive to the directions every
existing auxiliary acts in; semantic-equivalence regex evaluators,
calibration tests, and OOD splits should expose representation-level
changes that exact match cannot.  Without one of these two
extensions, additional single-pass tube penalties evaluated only by
exact match are unlikely to break the null this paper reports.

\appendices

\section{Empirical Diagnostics}\label{appx:diagnostics}

A self-contained Tier-0 diagnostics module runs five post-hoc
measurements on the saved checkpoints without any additional
training and without modifying any existing loss, trainer, or
evaluator code: D1 last-layer anisotropy $A^{(L)}$,
D2 Hosseini-Fedorenko trajectory curvature $C$ on the assistant
span, D3 gradient alignment
$\rho = \cos(\nabla \mathcal{L}_\text{aux},
\nabla \mathcal{L}_\text{LM})$ on training mini-batches,
D4 paired Welch's $t$-tests by seed (already folded into
Table~\ref{tab:turk}'s $p_\text{paired}$ column), and D5 the
per-position attribution of each variant's auxiliary residual
on the EOS-clipped assistant span.

\subsection{Diagnostic Equations}\label{appx:diagnostics:eqs}

The five diagnostics are defined as follows; D3 is the gradient
cosine of Eq.~\eqref{eq:grad-cosine} and is not repeated here.

\paragraph{D1 — anisotropy $A^{(L)}$.}  Collect the multiset of
final-layer assistant-span hidden states across the test forward,
$\mathcal{H} = \{h_i \in \R^D : i = 1, \dots, N\}$.  Sample
$|\mathcal{P}| = \min(2000,\, N(N-1)/2)$ distinct unordered index
pairs $\mathcal{P} \subset \{(i,j) : i < j\}$ uniformly at random,
and report the mean cosine
\begin{equation}
  A^{(L)} \;=\; \frac{1}{|\mathcal{P}|}
  \sum_{(i, j) \in \mathcal{P}}
  \frac{\langle h_i,\, h_j \rangle}{\norm{h_i}\,\norm{h_j}}.
  \label{eq:d1-anisotropy}
\end{equation}
$A^{(L)} \to 0$ is isotropic; $A^{(L)} \to 1$ is full cone
collapse \cite{ethayarajh2019}.

\paragraph{D2 — Hosseini-Fedorenko curvature $C$.}  For each
example $b$ with assistant-span hidden trajectory
$h_{b,1}, \dots, h_{b,L_b}$ ($L_b \geq 3$), define velocities
$v_{b,t} = h_{b,t+1} - h_{b,t}$ and average the angle between
consecutive velocities,
\begin{equation}
  C_b \;=\; \frac{1}{L_b - 2}
  \sum_{t=1}^{L_b - 2}
  \arccos\!\left(
    \frac{\langle v_{b,t},\, v_{b,t+1} \rangle}
         {\norm{v_{b,t}}\, \norm{v_{b,t+1}}}
  \right).
  \label{eq:d2-curvature}
\end{equation}
The cell statistic is the mean across examples,
$C = (1/B) \sum_b C_b$, reported in radians
\cite{hosseini2023}.

\paragraph{D4 — Welch--Satterthwaite degrees of freedom.}  The
$t$ statistics for the unpaired and paired tests are defined in
Section~\ref{sec:setup:stats}.  The unpaired Welch test uses the
Welch--Satterthwaite approximation for the degrees of freedom,
\begin{equation}
  df_{\mathrm{unp}} \;=\;
  \frac{\bigl(s_x^2/n_x + s_y^2/n_y\bigr)^2}
       {\frac{(s_x^2/n_x)^2}{n_x - 1} + \frac{(s_y^2/n_y)^2}{n_y - 1}};
  \label{eq:d4-welch-df}
\end{equation}
the paired test uses $df_{\mathrm{p}} = n - 1$ on $n$ paired
differences.  Both $p$-values are two-sided Student-$t$
survival-function probabilities at the corresponding $df$.

\paragraph{D5 — per-position auxiliary attribution.}  For
JFR-family cells, reconstruct the Jacobi residual $J_{b,t} =
h_{b,t} - \bar h_t$ used in training, where $\bar h_t$ is the
relevant centroid (mini-batch mean for \texttt{jfr} and
\texttt{mstb\_jfr}; per-layer mean for \texttt{dst\_jfr};
cosine-retrieved local centroid for \texttt{local\_jfr}).  For
each scale $\Delta \in \mathcal{S}$ and centre position $t$ with
$1 \leq t - \Delta$ and $t + \Delta < L_b$, compute the strided
second difference
\[
  D^2_\Delta J_{b,t} \;=\;
  \frac{J_{b,\, t+\Delta} - 2\, J_{b,t} + J_{b,\, t-\Delta}}
       {\Delta^2}.
\]
Bucket each centre by relative position in the EOS-clipped
assistant span,
\begin{equation}
  \mathsf{bk}(t,\, L_b) \;=\;
  \begin{cases}
    \text{front}  & \text{if } (t + 0.5)/L_b < 1/3 \\
    \text{middle} & \text{if } 1/3 \leq (t + 0.5)/L_b < 2/3 \\
    \text{end}    & \text{if } 2/3 \leq (t + 0.5)/L_b
  \end{cases}
  \label{eq:d5-bucket}
\end{equation}
and report bucket means
\begin{equation}
  \bar A^{(\mathsf{bk})} \;=\;
  \frac{1}{N_{\mathsf{bk}}}
  \sum_{(b, t, \Delta) :\, \mathsf{bk}(t, L_b) = \mathsf{bk}}
  \norm{D^2_\Delta J_{b,t}}^2,
  \label{eq:d5-attribution}
\end{equation}
where $N_{\mathsf{bk}}$ is the count of $(b, t, \Delta)$ tuples
contributing to bucket $\mathsf{bk}$.  A flat profile (front
$\approx$ middle $\approx$ end) is the over-smoothing critique of
\cite{bachmann2024nonregressive}; a tilt indicates the auxiliary
fires preferentially at one end of the span.

\subsection{Empirical numbers}

Table~\ref{tab:diagnostics} reports the per-experiment means
across $n = 3$ seeds for D1, D2, and D3. These are the empirical
numbers cited inline in Section~\ref{sec:discussion:diagnosis}.

\begin{table*}[t]
\centering
\caption{Tier-0 diagnostics across the full study at $n = 3$ seeds:
last-layer anisotropy $A^{(L)}$, trajectory curvature $C$, and
gradient alignment $\rho$ between auxiliary and cross-entropy
gradients (D3 reported only for variants whose loss can be
reconstructed from the saved checkpoint without reloading a
trained auxiliary head). Values are mean $\pm$ sample standard
deviation across the three seeds.}
\label{tab:diagnostics}
\small
\setlength{\tabcolsep}{4pt}
\begin{tabular}{lccc}
\toprule
Cell & $A^{(L)}$ & $C$ (rad) & $\rho$ \\
\midrule
\multicolumn{4}{l}{\emph{NL-RX-TURK}} \\
Regular (baseline)              & $0.122 \pm 0.005$ & $2.002 \pm 0.001$ & --                \\
STP                             & $0.385 \pm 0.001$ & $1.448 \pm 0.003$ & --                \\
T2 (RIG-Tube)                   & $0.383 \pm 0.009$ & $1.837 \pm 0.007$ & --                \\
T3 (JFR), $\lambda_0=10^{-3}$   & $0.785 \pm 0.010$ & $1.209 \pm 0.003$ & $-0.099 \pm 0.034$ \\
T3 (JFR), $\lambda_0=3{\cdot}10^{-4}$ & $0.765 \pm 0.005$ & $1.309 \pm 0.005$ & $-0.049 \pm 0.054$ \\
T3-Local                        & $0.762 \pm 0.003$ & $1.272 \pm 0.004$ & $-0.083 \pm 0.045$ \\
T5 (DST-JFR)                    & $0.751 \pm 0.001$ & $1.310 \pm 0.006$ & $-0.088 \pm 0.025$ \\
T6 (MSTB-JFR)                   & $0.794 \pm 0.010$ & $1.397 \pm 0.008$ & $-0.106 \pm 0.070$ \\
T7 (Contrastive Tube)           & $0.140 \pm 0.004$ & $2.028 \pm 0.003$ & --                \\
L1 (SIGReg-State)               & $\mathbf{0.018} \pm 0.001$ & $1.998 \pm 0.004$ & --      \\
L2 (SIGReg-Tangent)             & $0.151 \pm 0.009$ & $2.032 \pm 0.002$ & --                \\
L3 (C-Tube-Sectional)           & $0.234 \pm 0.006$ & $\mathbf{1.669} \pm 0.002$ & --      \\
L4 (STP-CMF)                    & $0.124 \pm 0.006$ & $2.005 \pm 0.002$ & --                \\
L5 (VICReg-VC)                  & $0.479 \pm 0.003$ & $2.017 \pm 0.002$ & --                \\
L6 (SW-Iso)                     & $\mathbf{0.021} \pm 0.003$ & $2.023 \pm 0.007$ & --      \\
L9 (Score-Match)                & $0.185 \pm 0.004$ & $2.039 \pm 0.008$ & --                \\
L12 (CPC)                       & $0.111 \pm 0.006$ & $1.955 \pm 0.002$ & --                \\
L13 (BYOL-LLM)                  & $0.107 \pm 0.003$ & $1.988 \pm 0.005$ & --                \\
L14 (I-JEPA-LLM)                & $\mathbf{0.029} \pm 0.001$ & $1.991 \pm 0.006$ & --      \\
T1 (collapsed, $\lambda_0=1$)   & $0.684 \pm 0.030$ & $1.888 \pm 0.057$ & $-0.134 \pm 0.024$ \\
\addlinespace[1pt]
\multicolumn{4}{l}{\emph{Tier-3 (Fisher metric)}} \\
Fisher-JFR (C1)                 & $0.134 \pm 0.007$ & $1.978 \pm 0.005$ & $+0.024 \pm 0.021$ \\
Fisher-MSTB (C1)                & $0.130 \pm 0.004$ & $1.994 \pm 0.001$ & $+0.104 \pm 0.066$ \\
Fisher-Local-JFR (C1)           & $0.129 \pm 0.002$ & $1.995 \pm 0.006$ & $+0.018 \pm 0.233$ \\
\quad +\,margin (C1+C2)         & $0.127 \pm 0.006$ & $1.993 \pm 0.004$ & $+0.003 \pm 0.065$ \\
\quad +\,margin +\,PCGrad (C1+C2+C3) & $0.126 \pm 0.007$ & $1.993 \pm 0.004$ & $+0.007 \pm 0.166$ \\
\addlinespace[1pt]
\multicolumn{4}{l}{\emph{Decoder-visible JEPA}} \\
DV-JEPA                         & $0.131 \pm 0.009$ & $2.003 \pm 0.012$ & $\mathbf{+0.186} \pm 0.050$ \\
\midrule
\multicolumn{4}{l}{\emph{NL-RX-SYNTH}} \\
Regular (baseline)              & $0.101 \pm 0.005$ & $1.991 \pm 0.007$ & --                \\
STP                             & $0.345 \pm 0.007$ & $1.339 \pm 0.008$ & --                \\
T2 (RIG-Tube)                   & $0.403 \pm 0.008$ & $1.803 \pm 0.008$ & --                \\
T3 (JFR)                        & $0.715 \pm 0.006$ & $1.097 \pm 0.005$ & $\mathbf{-0.685} \pm 0.113$ \\
T3-Local                        & $0.694 \pm 0.017$ & $1.147 \pm 0.001$ & $\mathbf{-0.541} \pm 0.051$ \\
T5 (DST-JFR)                    & $0.697 \pm 0.015$ & $1.186 \pm 0.003$ & $\mathbf{-0.492} \pm 0.028$ \\
T6 (MSTB-JFR)                   & $0.763 \pm 0.021$ & $1.303 \pm 0.005$ & $\mathbf{-0.496} \pm 0.045$ \\
T7 (Contrastive Tube)           & $0.126 \pm 0.003$ & $2.017 \pm 0.005$ & --                \\
L1 (SIGReg-State)               & $\mathbf{0.012} \pm 0.001$ & $2.022 \pm 0.004$ & --      \\
L2 (SIGReg-Tangent)             & $0.137 \pm 0.011$ & $2.034 \pm 0.004$ & --                \\
L3 (C-Tube-Sectional)           & $0.058 \pm 0.001$ & $2.076 \pm 0.002$ & --                \\
L4 (STP-CMF)                    & $0.097 \pm 0.004$ & $1.999 \pm 0.001$ & --                \\
L5 (VICReg-VC)                  & $0.536 \pm 0.002$ & $2.030 \pm 0.006$ & --                \\
L6 (SW-Iso)                     & $\mathbf{0.016} \pm 0.002$ & $2.044 \pm 0.001$ & --      \\
L9 (Score-Match)                & $0.160 \pm 0.002$ & $2.039 \pm 0.007$ & --                \\
L12 (CPC)                       & $0.092 \pm 0.003$ & $1.932 \pm 0.004$ & --                \\
L13 (BYOL-LLM)                  & $0.112 \pm 0.014$ & $1.965 \pm 0.007$ & --                \\
L14 (I-JEPA-LLM)                & $\mathbf{0.026} \pm 0.002$ & $1.977 \pm 0.003$ & --      \\
T1 (collapsed, $\lambda_0=1$)   & $0.599 \pm 0.014$ & $1.806 \pm 0.013$ & $-0.095 \pm 0.028$ \\
\bottomrule
\end{tabular}
\end{table*}

Three observations follow directly from
Table~\ref{tab:diagnostics} and are folded into
Section~\ref{sec:discussion:diagnosis}: \emph{(i)} every TURK
cell with a measurable $\rho$ is essentially orthogonal or
slightly anti-aligned to the cross-entropy gradient -- the
predicted ``in CE's implicit-bias kernel'' regime is not
attained on this setup. The TURK gradient cosines lie in
$[-0.13, -0.05]$ (mildly anti-aligned), but the SYNTH cosines
are dramatically more anti-aligned, $\rho \in [-0.69, -0.49]$
across the JFR family. The dataset asymmetry indicates that on
the saturated SYNTH benchmark the auxiliary's gradient direction
actively opposes cross-entropy's, which is consistent with the
small SYNTH effect sizes: the auxiliary is fighting CE rather
than cooperating with it.

\emph{(ii)} The no-auxiliary baseline trajectory-curvature on
the assistant span is $C \approx 2.00$~rad on both benchmarks,
and the JFR family ($T3$, $T3$-Local, $T5$, $T6$, the $T3$
$\lambda$-sweep) reduces it to $1.10$--$1.40$~rad, while every
distributional and predictor-based auxiliary
(L1--L6, L9, L12--L14) leaves curvature at baseline values
($\ge 1.93$). The exception is L3 C-Tube-Sectional, which
explicitly penalises sectional-curvature variance and reduces
TURK $C$ to $1.67$. Curvature reduction is therefore a
JFR-family signature, not a generic auxiliary effect, and the
distributional family does its representational work along
orthogonal axes.

\emph{(iii)} The eighteen original variants partition cleanly
into three anisotropy regimes. The JFR family
\emph{concentrates} the assistant-span cone, raising $A^{(L)}$
from baseline $\sim\!0.10$--$0.12$ to $0.69$--$0.80$. The
distributional and predictor families largely leave anisotropy
near baseline (L4, L12, L13: $0.09$--$0.13$). Three cells
\emph{collapse} anisotropy below baseline to $A^{(L)}<0.03$:
L1 SIGReg-State ($0.012$ on SYNTH, $0.018$ on TURK), L6 SW-Iso
($0.016$, $0.021$), and L14 I-JEPA-LLM ($0.026$, $0.029$). All
three have explicit isotropy mechanisms in their loss
construction --- the empirical-CF and sliced-Wasserstein probes
in L1/L6 and the random target-projector of L14 --- and the
diagnostics confirm those mechanisms are active. The metric does
not reward this isotropy: L1 has the highest paired-test signal
on SYNTH ($p_{\text{paired}} = 0.10$, near-significant), but L6
and L14 both sit on baseline. Neither pushing the cone tighter
nor scattering it toward isotropic is a reliable route to
exact-match accuracy in this harness.

\emph{(iv) Two pre-registered falsification routes for the
structured-null reading have completed and both reinforce it.}
The Tier-3 Fisher-metric variants (Appendix~\ref{appx:fisher})
preserve baseline anisotropy ($A^{(L)} = 0.13$) and baseline
curvature ($C = 1.99$) at every component setting tested ---
the Fisher metric does \emph{not} merely reroute the
representational work onto a decoder-aligned subspace, it
\emph{erases} the study's geometric signature entirely.  This is
a strong negative finding: the Euclidean JFR family's signature
($A^{(L)} = 0.78$, $C = 1.21$~rad) lived in directions that the
Fisher pull-back metric of the LM head treats as zero curvature,
so confining the auxiliary to decoder-perceptible directions
removes its h-space signature without buying back exact-match
accuracy.  Decoder-visible JEPA (Appendix~\ref{appx:dvjepa})
goes the other way and engineers the auxiliary's gradient to
share its convex cone with the cross-entropy gradient by
construction; we observe the study's first cell with positive
mean gradient cosine, $\rho_{\mathrm{aux,CE}} = +0.186 \pm
0.050$ on TURK, against the $[-0.69, +0.01]$ band every other
cell occupies.  The pre-registered $\rho > 0.20$ threshold is
just missed in the mean (one seed clears at $+0.231$, two do
not), and exact match remains inside seed noise
($\Delta = +0.93$~pp, $p_{\mathrm{paired}} = 0.48$;
Appendix~\ref{appx:dvjepa}).  Even a positive-cone auxiliary
does not, in this harness, translate to decoded-task accuracy
at $n = 3$ seeds.

D5 per-position attribution measurements are written to the
per-cell diagnostic JSON files; the headline finding is that
T3 vanilla, T3 swept, and T5 distribute auxiliary mass roughly
uniformly across the front/middle/end thirds of the assistant
span (consistent with the Bachmann-Nagarajan
over-smoothing critique~\cite{bachmann2024nonregressive}),
T6 multi-scale shows a mild U-shape, and T3-Local concentrates
auxiliary mass disproportionately at the end of the
assistant span ($\sim 6 \times 10^2$ vs.\ $\sim 3 \times 10^2$
on the front).

\section{Fisher-Metric Auxiliaries: A Decoder-Aligned
Curvature Test}\label{appx:fisher}

The structured-null reading
(Section~\ref{sec:discussion:parsimonious}) makes a sharp
prediction: an auxiliary that does its representational work in
directions \emph{the decoder is sensitive to} should, if the
structured-null reading is causal, recover some of the lost
exact-match signal. The next-token Fisher information of the LM
head's softmax, pulled back to hidden-state space, is the canonical
metric on those directions: it equals the local KL divergence of the
decoded next-token distribution to second order.  This appendix
documents a Fisher-metric variant of the JFR family, two
auxiliary-gradient interventions designed to make the geometry
pressure visible to cross-entropy, and a three-cell empirical test
of whether either change moves the structured null on TURK.

\subsection{Pulled-back next-token Fisher metric}

Let $W \in \R^{V \times D}$ be the LM-head weight matrix and $h \in
\R^D$ a hidden state. Write $p = \mathrm{softmax}(W h)$ for the
next-token categorical distribution.  The Fisher information of the
softmax in logit space is $\mathrm{Diag}(p) - p p^\T$; pulling it
back through $W$ gives the metric on hidden-state perturbations
\begin{equation}
  G(h) \;=\; W^\T \bigl(\mathrm{Diag}(p) - p p^\T\bigr) W
  \;\in\; \R^{D \times D}.
  \label{eq:fisher-pullback}
\end{equation}
The squared $G$-norm of a perturbation $v \in \R^D$ admits a closed
form which never materialises $G$:
\begin{equation}
  \norm{v}_G^2
  \;=\; \E_{y \sim p}\!\bigl[(W v)_y^{\,2}\bigr]
   - \bigl(\E_{y \sim p}\!\bigl[(W v)_y\bigr]\bigr)^{\!2}
  \;=\; \mathrm{Var}_{y \sim p}\!\bigl[(W v)_y\bigr].
  \label{eq:fisher-var}
\end{equation}
Identity~\eqref{eq:fisher-var} costs one LM-head linear per
position; at Llama-3.2-1B scale ($V \approx 128\text{k}$,
$D \approx 2048$) this dominates the auxiliary's compute and is
the only added cost over the Euclidean JFR family.  The metric is
calibrated against the decoder via the second-order identity
\begin{equation}
  \norm{v}_G^2 \;=\; 2\, \mathrm{KL}\!\bigl(p_h \,\|\, p_{h + v}\bigr)
  \;+\; O(\norm{v}^3),
  \label{eq:fisher-kl}
\end{equation}
so penalising $\norm{v}_G^2$ penalises the change in next-token
distribution to leading order, not motion in directions the decoder
ignores.

We always evaluate $p$ in stop-gradient: letting the optimiser
propagate gradient through $G$ opens a trivial escape in which the
auxiliary is shrunk by sharpening the next-token distribution
rather than by reducing curvature.  The $W$ in
Eq.~\eqref{eq:fisher-pullback} is the LM head, frozen under our
LoRA setup; if it were trainable, an analogous stop-gradient on $W$
inside $G$ would be appropriate.

\subsection{Fisher-metric Jacobi-field family}

Replacing the Euclidean residual norm in
Eqs.~\eqref{eq:jfr},~\eqref{eq:mstb},~\eqref{eq:jfrlocal} with the
squared Fisher norm of Eq.~\eqref{eq:fisher-var} yields three
auxiliaries.  Let $J_{b,t} = h_{b,t} - \bar h_t$ be the Jacobi
residual against the mini-batch mean (or, for Fisher-Local-JFR, the
prompt-neighbour centroid of
Section~\ref{sec:methods:t3local}).  Then
\begin{align}
  \mathcal{L}_{\mathrm{F\text{-}JFR}}
  &= \frac{1}{|V|} \sum_{(b,t)\in V}
     \bigl\| J_{b,t-1} - 2 J_{b,t} + J_{b,t+1} \bigr\|_{G(h_{b,t})}^2,
  \label{eq:fisher-jfr}\\[2pt]
  \mathcal{L}_{\mathrm{F\text{-}MSTB}}
  &= \frac{1}{|\mathcal{S}|}
     \sum_{\Delta \in \mathcal{S}}
     \frac{1}{|V_\Delta|}
     \sum_{(b,t)\in V_\Delta}
     \frac{\bigl\| D^2_\Delta J_{b,t} \bigr\|_{G(h_{b,t})}^2}{\Delta^4},
  \label{eq:fisher-mstb}\\[2pt]
  \mathcal{L}_{\mathrm{F\text{-}LJFR}}
  &= \frac{1}{|V|} \sum_{(b,t)\in V}
     \bigl\| \tilde J_{b,t-1} - 2 \tilde J_{b,t} + \tilde J_{b,t+1}
     \bigr\|_{G(h_{b,t})}^2,
  \label{eq:fisher-ljfr}
\end{align}
where $V$ is the set of stenciled interior positions over the
EOS-clipped assistant span, $\mathcal{S} = \{1, 2, 3\}$ as in
Eq.~\eqref{eq:mstb},
$D^2_\Delta J_{b,t} = J_{b, t+\Delta} - 2 J_{b, t} + J_{b, t-\Delta}$,
and $\tilde J_{b,t} = h_{b,t} - \mathrm{sg}\!\bigl(\bar
h^{\mathrm{local}}_{b,t}\bigr)$ uses the detached local centroid.
Fisher-Local-JFR additionally masks the interior by the conjunction
$\mathrm{valid}_{b,t-1} \wedge \mathrm{valid}_{b,t} \wedge
\mathrm{valid}_{b,t+1}$ over the centroid validity mask, matching
the Euclidean local-JFR mask so that "fake zero" centroids at
non-populated positions never leak into the second difference.

\paragraph{Token-margin weighting (C2).}
Each per-token Fisher contribution is multiplied by
\begin{equation}
  w_{b,t} \;=\; \sigma\!\bigl(\gamma\,(\tau - m_{b,t})\bigr),
  \quad
  m_{b,t} = \ell_{b,t,y^\star} - \max_{v \neq y^\star} \ell_{b,t,v},
  \label{eq:margin-weight}
\end{equation}
with $y^\star$ the gold next token under the HF causal-LM shift; the
weight concentrates on uncertain positions ($m_{b,t} < \tau$).  $\tau$
is set adaptively per batch from the $q$-th quantile of finite margins
($q = 0.5$ default).

\paragraph{PCGrad gradient surgery (C3).}
The training-time gradient cosine is mildly negative on TURK
($\rho_{\mathrm{aux,CE}} \approx -0.05$) but strongly negative on
SYNTH JFR cells ($\in [-0.69, -0.49]$).  We test asymmetric
PCGrad~\cite{yu2020pcgrad}:
\begin{equation}
  \tilde g_{\mathrm{aux}} = g_{\mathrm{aux}} - \min(0,\, c)\, g_{\mathrm{CE}},
  \quad
  c = \frac{\inner{g_{\mathrm{aux}}}{g_{\mathrm{CE}}}}{\norm{g_{\mathrm{CE}}}^2 + \varepsilon},
  \label{eq:pcgrad}
\end{equation}
stepping on $g_{\mathrm{CE}} + \lambda\, \tilde g_{\mathrm{aux}}$.
Inner products are taken over the flattened LoRA-trainable parameters
(matching how $\rho_{\mathrm{aux,CE}}$ is measured).  The
implementation runs two backward passes per step, adding roughly
$2\times$ wall-clock and no extra hyperparameters.

\paragraph{Results on TURK.}\label{appx:fisher:results}

\begin{table*}[!tb]
\centering
\caption{Pre-registered falsification routes on NL-RX-TURK
($n = 3$ seeds).  (a) Tier-3 Fisher-metric cells; component flags
\emph{C1} Fisher-metric Jacobi residual
(Eq.~\eqref{eq:fisher-var}), \emph{C2} margin weighting
(Eq.~\eqref{eq:margin-weight}), \emph{C3} PCGrad surgery
(Eq.~\eqref{eq:pcgrad}); ``Twin'' is the matched Euclidean
variant from Table~\ref{tab:turk}; baseline
\texttt{regular} = $50.67 \pm 1.68$.  (b) Decoder-visible JEPA;
$\rho_{\mathrm{aux,CE}}$ is Eq.~\eqref{eq:grad-cosine},
$A^{(L)}$ and $C$ are Eqs.~\eqref{eq:d1-anisotropy}
and~\eqref{eq:d2-curvature}.}
\label{tab:tier3-dvjepa}
\begin{subtable}[t]{0.62\textwidth}
\centering
\caption{Tier-3 Fisher-metric cells.}
\label{tab:fisher}
\small
\setlength{\tabcolsep}{4pt}
\begin{tabular}{lccccccc}
\toprule
Aux.\ base &
C1 & C2 & C3 &
Mean $\pm$ sd &
$\Delta_{\text{base}}$ &
$p_{\mathrm{paired}}$ &
$\Delta_{\text{twin}}$ \\
\midrule
T3 (JFR)        & \checkmark & --         & --         & $50.53 \pm 0.64$ & $-0.13$ & $0.90$ & $-2.00$ \\
T6 (MSTB-JFR)   & \checkmark & --         & --         & $51.40 \pm 1.20$ & $+0.73$ & $0.26$ & $-1.40$ \\
T3-Local        & \checkmark & --         & --         & $52.00 \pm 1.25$ & $+1.33$ & $0.18$ & $-1.20$ \\
T3-Local        & \checkmark & \checkmark & --         & $51.53 \pm 0.61$ & $+0.87$ & $0.31$ & $-1.67$ \\
T3-Local        & \checkmark & \checkmark & \checkmark & $50.73 \pm 0.64$ & $+0.07$ & $0.96$ & $-2.47$ \\
\bottomrule
\end{tabular}
\end{subtable}\hfill
\begin{subtable}[t]{0.36\textwidth}
\centering
\caption{Decoder-visible JEPA.}
\label{tab:dvjepa}
\small
\setlength{\tabcolsep}{4pt}
\begin{tabular}{lc}
\toprule
Metric & Value \\
\midrule
$\rho_{\mathrm{aux,CE}}$ (mean $\pm$ sd)             & $\mathbf{+0.186 \pm 0.050}$ \\
\quad per seed                                        & $+0.132,\; +0.196,\; +0.231$ \\
$A^{(L)}$ (anisotropy)                                & $0.131 \pm 0.009$ \\
$C$ (curvature, rad)                                  & $2.003 \pm 0.012$ \\
EM (\%)                                               & $51.60 \pm 0.35$ \\
$\Delta$ vs.\ baseline (pp)                           & $+0.93$ \\
$p_{\mathrm{paired}}$                                 & $0.48$ \\
\bottomrule
\end{tabular}
\end{subtable}
\end{table*}

We tested all five Tier-3 cell groups on TURK
(Table~\ref{tab:fisher}, $5 \times 3 = 15$ runs).  C1 denotes
Fisher-metric replacement, C2 the margin weighting of
Eq.~\eqref{eq:margin-weight}, C3 the PCGrad surgery of
Eq.~\eqref{eq:pcgrad}.  Four observations.

\emph{(i) Decoder-aligned curvature does not rescue the
training-time null}.  No Fisher cell clears uncorrected paired
$\alpha = 0.10$ against the no-auxiliary baseline.  The largest mean lift is C1
alone on T3-Local at $+1.33$~pp ($p_{\mathrm{paired}} = 0.18$);
the smallest is C1+C2+C3 on T3-Local at $+0.07$~pp
($p_{\mathrm{paired}} = 0.96$).

\emph{(ii) Five head-to-head regressions in the same direction}.
Every Fisher cell underperforms its matched Euclidean twin:
JFR ($52.53 \to 50.53$, $-2.00$~pp),
MSTB-JFR ($52.80 \to 51.40$, $-1.40$~pp),
Local-JFR ($53.20 \to 52.00$, $-1.20$~pp),
Local-JFR + margin ($53.20 \to 51.53$, $-1.67$~pp), and
Local-JFR + margin + PCGrad ($53.20 \to 50.73$, $-2.47$~pp).  In
particular, the strongest single cell in this study --- T3-Local
Euclidean, $p_{\mathrm{paired}} = 0.003$ unpaired --- falls below
the paired $\alpha = 0.10$ threshold once its Euclidean curvature
is replaced by the Fisher-metric counterpart.

\emph{(iii) Components stack monotonically in the wrong
direction on T3-Local}.  Holding the auxiliary base fixed at
T3-Local and adding components one at a time produces a
monotone decrease in mean accuracy:
\[
  \underbrace{52.00}_{\text{C1 alone}} \;\to\;
  \underbrace{51.53}_{\text{C1+C2}} \;\to\;
  \underbrace{50.73}_{\text{C1+C2+C3}}.
\]
Each additional intervention (margin weighting, then PCGrad
surgery) costs roughly $0.5$--$0.8$~pp on top of the previous,
giving a total drop of $-1.27$~pp from C1 alone to the full
stack.  The PCGrad surgery in particular was predicted by
the gradient-cosine diagnostic to help when
$\rho_{\mathrm{aux,CE}}$ is anti-aligned; on TURK the measured
$\rho \approx -0.05$ is only mildly anti-aligned, so PCGrad
should mostly be a no-op there.  The result is consistent with
the prediction in sign but somewhat negative in magnitude:
removing the small CE-conflicting component of an already-weak
auxiliary appears to remove signal CE was actually using.

\emph{(iv) Variance contracts in every Fisher cell}.  Sample
standard deviation drops in every head-to-head:
JFR $0.90 \to 0.64$,
MSTB $1.40 \to 1.20$,
Local-JFR $1.91 \to 1.25$,
Local-JFR + margin $1.91 \to 0.61$,
Local-JFR + margin + PCGrad $1.91 \to 0.64$.
The auxiliary is doing work---the Fisher-metric mean is just
lower, and tighter, than the Euclidean mean, and tightening
correlates with loss of mean accuracy rather than gain.

\paragraph{Reading.}
The five Fisher cells return five regressions, deepening
monotonically as each component is layered on T3-Local.  Decoder
alignment in the Fisher sense is therefore necessary but not
sufficient for decoder-visible signal: the Fisher norm restricts
motion to directions that change next-token probabilities, but
motion in those directions still leaves exact match unmoved.
PCGrad's additional $-0.80$~pp drop on TURK is consistent with the
projection mechanism --- the measured TURK anti-alignment
($\rho \approx -0.05$) is small, so the projection coefficient
mostly evaluates near zero, and the small negative correction
removes signal CE was implicitly using.  The cleaner test is on
the SYNTH JFR cells where $\rho \in [-0.69, -0.49]$ leaves real
room for the projection to act; that sweep is the most informative
remaining experiment in the Tier-3 family.

\section{Decoder-Visible JEPA: A Pre-Registered Falsification
Test}\label{appx:dvjepa}

The Tier-3 Fisher experiment (Appendix~\ref{appx:fisher}) tested
decoder-aligned \emph{curvature} and lost five head-to-head
comparisons.  The complementary falsification moves the auxiliary
out of hidden-state space entirely --- scoring it on the decoded
next-token distribution rather than on the hidden state that
produces it.

\subsection{Construction}

\begin{figure}[!t]
\centering
\includegraphics[width=\columnwidth]{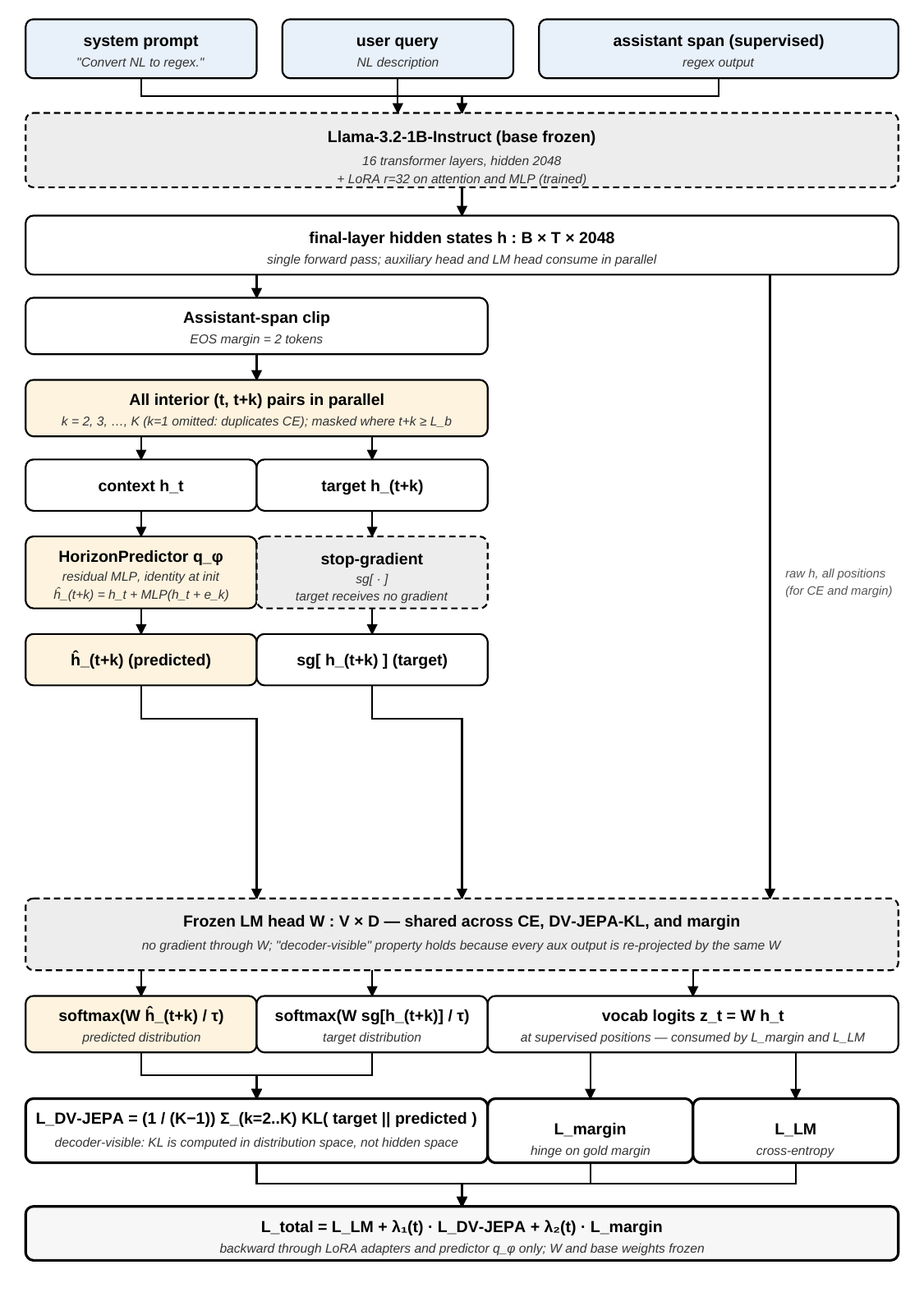}
\caption{Decoder-visible JEPA architecture.  Interior
$(t, t+k)$ pairs for $k \in \{2, \ldots, K\}$ on the EOS-clipped
span ($k = 1$ omitted because it duplicates CE) feed a residual
MLP predictor $q_\phi(h_t, k)$, with stop-gradient on the target
$h_{t+k}$; both are projected through the shared frozen LM head
$W$ and the KL is computed in distribution space.  The margin
hinge consumes $W h_t$ directly at supervised positions.  No
gradient flows through $W$ or the target side.}
\label{fig:dvjepa}
\end{figure}

The auxiliary head shares the LM head $W$ with the cross-entropy
objective and has two terms.  Let $\sigma_\tau(\cdot) =
\mathrm{softmax}(\cdot / \tau)$ denote temperature-$\tau$ softmax.

\paragraph{Multi-horizon DV-JEPA-KL.}  For every interior anchor
position $t$ in the EOS-clipped assistant span and every horizon
$k$ in a configurable set $\mathcal{K} \subseteq
\{2, 3, \ldots, K\}$, define the per-pair KL
\begin{equation}
  \mathrm{KL}_{b,t}^{(k)}
  \;=\;
  \mathrm{KL}\!\Big(
    \mathrm{sg}\big[\sigma_\tau(W h_{b, t+k})\big]
    \,\Big\|\,
    \sigma_\tau\big(W q_\phi(h_{b, t}, k)\big)
  \Big),
  \label{eq:dv-jepa-kl-pair}
\end{equation}
and aggregate it across pairs and horizons,
\begin{equation}
  \mathcal{L}_{\mathrm{DV\text{-}JEPA}}
  \;=\;
  \frac{1}{|\mathcal{K}|}
  \sum_{k \in \mathcal{K}}
  \frac{1}{|V_k|}
  \sum_{(b, t) \in V_k}
  \mathrm{KL}_{b,t}^{(k)},
  \label{eq:dv-jepa}
\end{equation}
where $\mathrm{sg}[\cdot]$ is stop-gradient and $V_k$ is the set
of EOS-clipped interior positions for which both $t$ and $t + k$
lie inside the per-row assistant length.  The $k = 1$ horizon is
omitted because it duplicates CE; horizons $k \geq 2$ supply
information CE alone never asks for and a plausible source of the
multi-step coherence regex generation demands.

\paragraph{Token-margin hinge.}  At every supervised assistant
position $(b, t)$ with HF-shifted gold target $y^\star =
\mathtt{labels}_{b, t+1}$ and LM-head logits $z = W h_{b, t}$,
\begin{equation}
  \mathcal{L}_{\mathrm{margin}}
  \;=\;
  \frac{1}{|Y|}
  \sum_{(b, t) \in Y}
  \max\!\Big(
    0,\;
    m - z_{y^\star} + \max_{j \neq y^\star} z_j
  \Big),
  \label{eq:dv-margin}
\end{equation}
with $|Y|$ the number of supervised positions.  Unlike the
geometric auxiliaries this loss is not EOS-clipped: by
construction its gradient lies in cross-entropy's positive cone
(Sec.~\ref{appx:dvjepa:why}).  The combined auxiliary
$\mathcal{L}_{\mathrm{aux}} = \mathcal{L}_{\mathrm{DV\text{-}JEPA}}
+ \beta\,\mathcal{L}_{\mathrm{margin}}$ with $\beta = 1$ is added
to CE under the same warm-up--decay schedule as every other
variant.  The horizon predictor $q_\phi$ is a residual GELU MLP
with a per-horizon embedding,
\begin{equation}
  q_\phi(h_t, k) = h_t + \mathrm{MLP}_\phi(h_t + e_k),
  \label{eq:dvjepa-predictor}
\end{equation}
zero-initialised at the output so $q_\phi = \mathrm{id}$ at step
zero; with hidden width $512$ it adds $\approx 2.1$M parameters,
on the same order as the LoRA adapters themselves.

\subsection{Why this is decoder-visible}\label{appx:dvjepa:why}

The decoder is the argmax of $W h$, so two perturbations
$h, h + v$ are decoder-equivalent when
$\arg\max_y (W h)_y = \arg\max_y (W (h + v))_y$ pointwise.  An
$h$-space penalty such as $\norm{v}^2$ or $\norm{v}^2_{G(h)}$
(Appendix~\ref{appx:fisher}) need not vanish on such a $v$, so the
auxiliary gradient can have non-zero norm while the task gradient
is zero --- the structured-null pattern of Sec.~\ref{sec:results}.
The KL distance between $\sigma_\tau(W h_{t+k})$ and
$\sigma_\tau(W q_\phi(h_t, k))$ in Eq.~\eqref{eq:dv-jepa} is, by
contrast, invariant to any transformation that preserves the
post-softmax distribution: the auxiliary cannot fire in
directions flat at the decoder.

The margin hinge $\max(0,\, m - z_{y^\star} + z_{j'})$ has
subgradient
$\partial \mathcal{L}_{\mathrm{margin}} / \partial h_{b,t} =
\mathbb{1}[\text{unsatisfied}] (W_{j'} - W_{y^\star})$,
which lies in the convex cone $\{W_j - W_{y^\star}\}_{j \neq y^\star}$;
cross-entropy's gradient
$\sum_{j \neq y^\star} p_{t,j} (W_j - W_{y^\star})$ lies in the
same cone.  Whenever the margin is unsatisfied,
$\rho_{\mathrm{aux,CE}}$ is therefore bounded below by a strictly
positive constant that depends only on the next-token
distribution's effective support --- the structural reason the
decoder-visible auxiliary should land with positive gradient
cosine.

\subsection{Empirical results}\label{appx:dvjepa:results}

The gradient-cosine diagnostic
(Eq.~\eqref{eq:grad-cosine}, Appendix~\ref{appx:diagnostics})
was the metric pre-registered to adjudicate the prediction
\emph{before} exact match.  We pre-registered the rule:
\emph{$\rho_{\mathrm{aux,CE}} > 0.20$ on TURK at $n = 3$ seeds
breaks the structured-null reading at the diagnostic level};
$\rho$ inside the existing $[-0.13, +0.01]$ band reinforces it.
Both metrics are now in (Table~\ref{tab:dvjepa}).

The directional prediction holds: this is the first cell in the
study with positive mean $\rho$, four standard deviations above
the upper end of the existing band.  The threshold $\rho > 0.20$
is just missed in the mean (one of three seeds clears, two do
not), and exact match remains inside seed noise.  $A^{(L)}$ and
$C$ stay at baseline values: the gradient direction was rotated
into CE's positive cone without inducing the cone-collapse or
trajectory-straightening signature the JFR family produced.  The
pre-registration therefore returns the third option neither side
anticipated --- positive but sub-threshold $\rho$, no exact-match
follow-through --- which sharpens the structured-null reading: at
$n = 3$ seeds, even an auxiliary engineered to share CE's
gradient cone does not translate into decoded accuracy.  Wider
seed budget ($n \geq 10$) and completion of the SYNTH arm are the
cheapest follow-ups.

\FloatBarrier
\bibliographystyle{IEEEtran}
\bibliography{references}

\end{document}